\documentclass[10pt,journal,compsoc]{IEEEtran}
\usepackage{hyperref}

\newcommand{\subbold}[1]{\vspace{0.1cm}\noindent\textbf{#1}\;}
\usepackage{multicol}

\usepackage{makecell}

\usepackage{graphics} 
\usepackage{epsfig} 
\usepackage{amsmath} 
\usepackage{amssymb}
\usepackage[dvipsnames]{xcolor}
\definecolor{peach}{RGB}{255,220,102}

\usepackage{transparent}

\usepackage{subcaption}
\usepackage{enumitem}

\usepackage{graphicx}
\input{insbox}

\usepackage{booktabs}

\usepackage{caption, copyrightbox}
 
\newcommand{\cut}[1]{}

\newcommand{\mj}[1]{\textcolor{black}{#1}}
\newcommand{\mcorr}[1]{\textcolor{black}{#1}}

\usepackage[numbers,sort&compress,square]{natbib}

\usepackage{wrapfig}

\usepackage{lipsum}
\usepackage{tcolorbox}

\begin{document}


\title{Survey: Leakage and Privacy at Inference Time}
%
%
%

\author{Marija Jegorova,
        Chaitanya Kaul,
        Charlie Mayor,
        Alison~Q.~O'Neil,
        Alexander~Weir,
        Roderick Murray-Smith,
        and~Sotirios~A.~Tsaftaris
\thanks{M. Jegorova is now with Facebook FAIAR, London but work was completed when M. Jegorova was with the University of Edinburgh, UK. C. Kaul and R. Murray-Smith are with University of Glasgow. C. Mayor is with NHS Scotland. A. O'Neil and A. Weir are with Canon Medical Research Europe, Edinburgh, UK. S.A. Tsaftaris is with the University of Edinburgh, UK and The Alan Turing Institute, London, UK.}
}



\IEEEtitleabstractindextext{%
\begin{abstract}

\mj{Leakage of data from publicly available Machine Learning (ML) models is an area of growing significance since commercial and government applications of ML can draw on multiple sources of data, potentially including users’ and clients’ sensitive data. We provide a comprehensive survey of }%
%
contemporary advances on several fronts, covering involuntary data leakage which is natural to ML models, potential malicious leakage which is caused by privacy attacks, and currently available defence mechanisms. We focus on inference-time leakage, as the most likely scenario for publicly available models. We first discuss what leakage is in the context of different data, tasks, and model architectures. We then propose a taxonomy across involuntary and malicious leakage, followed by description of currently available defences, assessment metrics, and applications. We conclude with outstanding challenges and open questions, outlining some promising directions for future research.

\end{abstract}

}

\maketitle



\IEEEpeerreviewmaketitle

\vspace{-1cm}

\section{Introduction}

Machine Learning (ML) technologies have become prolific in modern day life, with many ML models made publicly available. Data leakage is an area of growing significance as commercial and government applications of ML can draw on multiple sources of data, potentially including users’ and clients’ sensitive data. Hence, it is important to understand the potential leakage scenarios and existing prevention mechanisms in order to safeguard against revealing information about models' training data, in particular data which breaches an individual's privacy.



\begin{figure}
    \centering
    \includegraphics[scale=0.65]{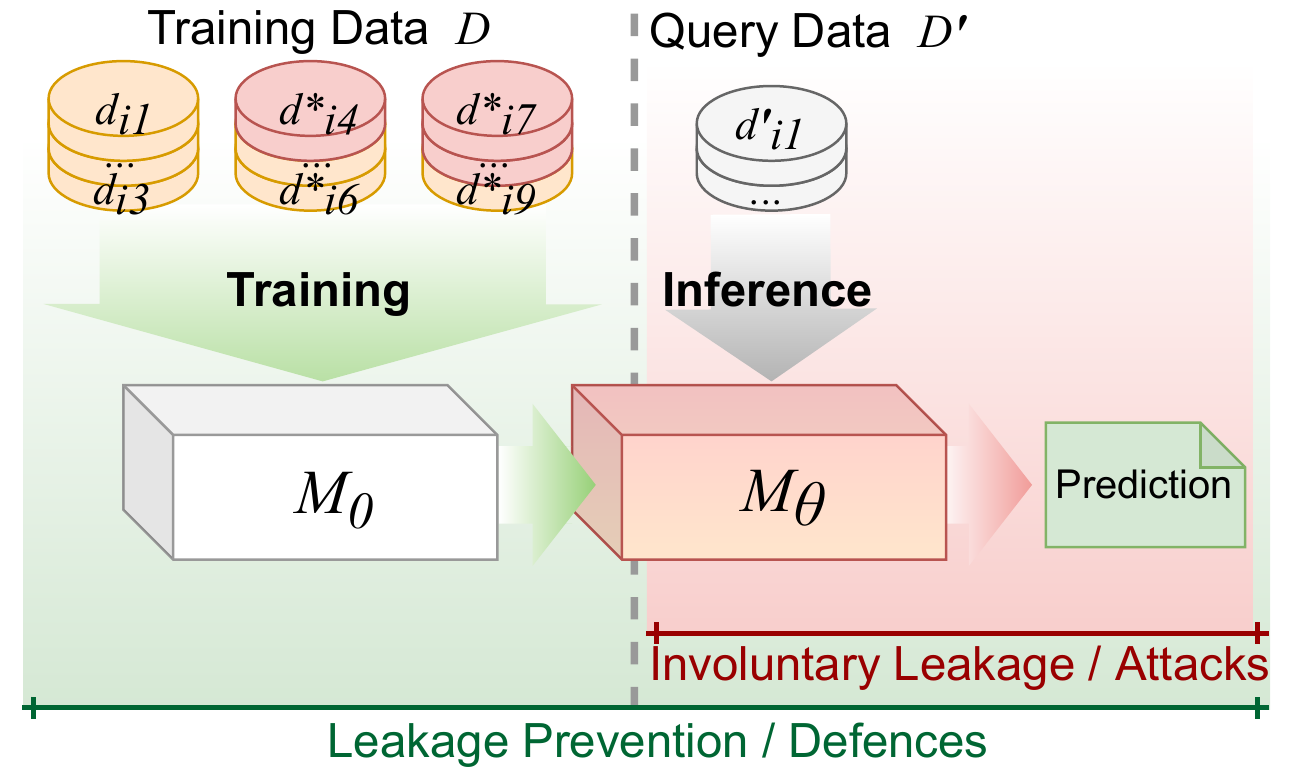}
    
    \caption{We survey benign leakage and attacks at inference time, concerning the query dataset $D'$, trained model $M_\theta$, parameters $\theta$, and predictions (or confidence scores) of $M_\theta$.
    }
    \label{fig:broad_scope}
    \vspace{-0.4cm}
\end{figure}

To address this need, we present a comprehensive overview and unified perspective on data leakage in trained ML models, including: causes of involuntary leakage, the implications of these causes being exploited by malevolent users, the methods for measuring and preventing such attacks, and finally the challenges and opportunities for further research into data leakage. To the best of our knowledge, existing surveys on privacy focus on privacy attacks or some subset of them \cite{Survey_Biggio2018WildPT, Survey_towardsAdversarialMalwareDetection, Survey_theSecurityOfMLinAdversarialSetting, Survey_SOK, Survey_OverviewOnPrivacy_ArXiv, Survey_ofPrivacyAttacks, Chang2018PrivacyInNN_ThreatsAndCountermeasures, Survey_DeepLearningForPrivacy_Tanuwidjaja2019ASO, Survey_small_Zhang2020PrivacyTA, Survey_Rezaei2019SecurityOD, Survey_whenMLmeetsPrivacy}, whereas we examine the broader picture of data leakage. Since the interest of this survey lies primarily in data leakage from trained models, we review research focused on inference time leakage and attacks (see Figure~\ref{fig:broad_scope}). If the reader would like to examine training time interventions, there are a number of relevant surveys \cite{Survey_AdversarialAttacksAndDefencesOZDAG2018152, Survey_AdversarialAttackDefences_2018, Survey_AdversarialAttacksAndDefences_inImagesGraphsText, Survey_Adversarial_2020_REN2020346, Survey_SOK}. 
Our contributions are as follows:
\begin{itemize}
    \item \textit{first comprehensive survey on data leakage}, including involuntary and malicious leakage methodology, prevention and defences, assessment metrics, and applications;
    \item acknowledging that leakage is context-specific, 
    we describe the data leakage research conducted in different task and data type contexts;
    \item in-depth presentation of current methodologies; and
    \item a summary of the challenges and open questions in the data leakage research field.
\end{itemize}

The paper is structured as follows: Section~\ref{sec:definitions} provides definitions and notation, and discusses what private and sensitive data are in 
a variety of contexts. Section~\ref{sec:leakage} covers causes of natural \textit{involuntary} data leakage, whilst Section~\ref{sec:attacks} covers \textit{privacy attacks}. Sections~\ref{sec:defences_data} and \ref{sec:defences_model} cover leakage prevention and defence mechanisms. Sections~\ref{sec:metrics} provides an aggregated picture of currently available metrics for assessing leakage and privacy. Section~\ref{sec:apps} outlines applications
. We conclude with remaining challenges and open questions in Sections~\ref{sec:challenges} and \ref{sec:conclusion}.

\section{Definitions}
\label{sec:definitions}

First of all let us define the notation for this article. Every ML model $M_{\theta}$, regardless of its task, is trained on some data $D$, which consists of the individual data samples $d_i$ which have features $f_j$, some of which are sensitive $f^{*}_j$ ($i = 1, ...k$, where $k$ is the number of data samples in $D$, and $j = 1, ..., n$, where $n$ is number of features of $D$). With respect to this notation:
\begin{itemize}
    \item Data leakage, differential privacy, membership inference attacks, and data reconstruction attacks, are all focused around the safety of the individual data samples $d_i$, i.e. the possibility of inferring 
    $d_i$ from the model $M_{\theta}$.
    \item Feature leakage and property inference attacks are concerned with inferring some properties of the sensitive features $f^{*}_j$ of the training dataset $D$.
    \item Model extraction attacks are interested in inferring the parameters $\theta$ of the trained model $M_{\theta}$ (or their feasible approximations) in order to steal this model.
\end{itemize}

The end-user can have different levels of access to the trained model $M_\theta$. Traditionally, these are separated into black- and white-box access. \textbf{Black-box access}, or \textbf{query access}, assumes that the user controls the input and has access to the output of $M_\theta$. \textbf{White-box access} assumes that the user has full access to $M_\theta$, its input, output, architecture, and parameters $\theta$.
\textbf{Gray-box access} describes situations in between, e.g. user might not know the model's architecture and parameters $\theta$ but has access to outputs from the model's intermediate layers, or might not know the parameters $\theta$, but has access to the architecture of the model $M_\theta$, \textit{etc}.
  

\subsection{What is personal (private) and sensitive data?}

We distinguish between \textit{personal}, \textit{personal ``sensitive''}, and \textit{non-personal} data.


\begin{table*}[htbp]
\setlength{\tabcolsep}{3pt}
\renewcommand{\arraystretch}{1.25}
\begin{center}
\begin{tabular}{l | l l l l || l l l l |l}
\hline
\textbf{Type of Data Leakage} & \multicolumn{4}{c||}{\textbf{Type of Data}} & \multicolumn{5}{c}{\textbf{Type of Tasks}} \\
\cline{2-10}
 & Images & Text & Tabular & Time-series & Classification & Regression & Generation & Seg.& MLaaS\\
\hline
\hline
\hspace{-3pt}\textbf{Involuntary leakage:}& & & & & & & & \\
Overfitting Sec. \ref{sec:overfitting}& \cite{Memorisation_Generalisation_ICLR2017,Overfitting_PrivacyRiskInML_AnalysingConnectionToOverfitting,Overfitting_UnintendedConsequencesOfOverfitting,MIA_Nasr2019ComprehensivePA,UnderstandingMI_onWellGeneralisedML} & - & \cite{Overfitting_PrivacyRiskInML_AnalysingConnectionToOverfitting,Overfitting_UnintendedConsequencesOfOverfitting,MIA_Nasr2019ComprehensivePA,UnderstandingMI_onWellGeneralisedML, Leaks_Song2020OverlearningRS}&-& \cite{Memorisation_Generalisation_ICLR2017,Overfitting_PrivacyRiskInML_AnalysingConnectionToOverfitting,Overfitting_UnintendedConsequencesOfOverfitting,MIA_Nasr2019ComprehensivePA,UnderstandingMI_onWellGeneralisedML}& \cite{Overfitting_PrivacyRiskInML_AnalysingConnectionToOverfitting,Overfitting_UnintendedConsequencesOfOverfitting, UnderstandingMI_onWellGeneralisedML}&-&-&-\\

Memorization Sec. \ref{sec:mem}& \cite{Memorisation_MLmodelsThatRememberTooMuch, DEJAVU, MemorisationPrecedesGeneration}& \cite{TheSecretSharer_EvaluatingUnintendedMemorisation,Memorisation_MLmodelsThatRememberTooMuch,TheSecretSharer_MeasuringMemorisation_and_ExtractingSecrets}& \cite{MemGANsWorkshop}& - & \cite{Memorisation_MLmodelsThatRememberTooMuch, DEJAVU}& - & \cite{TheSecretSharer_EvaluatingUnintendedMemorisation,TheSecretSharer_MeasuringMemorisation_and_ExtractingSecrets, MemGANsWorkshop,MemorisationPrecedesGeneration}& -&-\\

Feature Leakage Sec. \ref{sec:featureLeak}& \cite{F-BLEAU-fast-black-box-leakage-estimation,ExploitingUnintendedFeatureLeakage_inCollaborativeLearning,Leaks_Song2020OverlearningRS}&  \cite{TheSecretSharer_EvaluatingUnintendedMemorisation, TheSecretSharer_MeasuringMemorisation_and_ExtractingSecrets, ExploitingUnintendedFeatureLeakage_inCollaborativeLearning} & \cite{F-BLEAU-fast-black-box-leakage-estimation,ExploitingUnintendedFeatureLeakage_inCollaborativeLearning,Leaks_Song2020OverlearningRS}& - & \cite{F-BLEAU-fast-black-box-leakage-estimation,ExploitingUnintendedFeatureLeakage_inCollaborativeLearning,Leaks_Song2020OverlearningRS}& \cite{F-BLEAU-fast-black-box-leakage-estimation, Leaks_Song2020OverlearningRS}& \cite{TheSecretSharer_EvaluatingUnintendedMemorisation, TheSecretSharer_MeasuringMemorisation_and_ExtractingSecrets,ExploitingUnintendedFeatureLeakage_inCollaborativeLearning}& - &-\\

\hline
\hline

\hspace{-3pt}\textbf{Malicious Leakage / Attacks:}& & & & & & & & \\
Membership Infer. Sec. \ref{subsec:MIAs}&\makecell[l]{\cite{Overfitting_PrivacyRiskInML_AnalysingConnectionToOverfitting, Overfitting_UnintendedConsequencesOfOverfitting,MIA_Nasr2019ComprehensivePA,UnderstandingMI_onWellGeneralisedML} \\  \cite{GAN-Leaks, MIA_Salem2019MLLeaksMA, MIA_DemystifyingMIAsMLasService,MIA_ML, MIA_Rahman2018MembershipIA_againstDifferentiallyPrivate,MIA_PrivacyRisks_of_SecuringML_againstAdversarialExamples, MIA_DeepModelsUnderGAN, sablayrolles19a_BayesOptimalStrategiesforMIA,LOGAN_v1,LOGAN_v2}}& 
\cite{MIA_Liu2019SocInfMI, MIA_hisamoto-etal-2020-membership_MachineTranslation, ExploitingUnintendedFeatureLeakage_inCollaborativeLearning} &\makecell[l]{
\cite{ExploitingUnintendedFeatureLeakage_inCollaborativeLearning,Overfitting_PrivacyRiskInML_AnalysingConnectionToOverfitting,Overfitting_UnintendedConsequencesOfOverfitting,MIA_Nasr2019ComprehensivePA,UnderstandingMI_onWellGeneralisedML}\\ \cite{MIA_Salem2019MLLeaksMA,MIA_DemystifyingMIAsMLasService, ReID_usingGenerativeModels, Reconstruction_PrivacyInPharmcogenetics, MIA_Long2017TowardsMeasuringMembershipPrivacy}}&\makecell[l]{
\cite{Reconstruction_PrivacyInPharmcogenetics,UnderTheHoodOfMIAonAggregateLocationTS} \\ \cite{MIA_OnAggregateLocationData}}&
\makecell[l]{
\cite{ExploitingUnintendedFeatureLeakage_inCollaborativeLearning,Overfitting_PrivacyRiskInML_AnalysingConnectionToOverfitting,Overfitting_UnintendedConsequencesOfOverfitting,MIA_Nasr2019ComprehensivePA,UnderstandingMI_onWellGeneralisedML}\\
\cite{sablayrolles19a_BayesOptimalStrategiesforMIA,MIA_Salem2019MLLeaksMA,MIA_DemystifyingMIAsMLasService,MIA_ML, MIA_Rahman2018MembershipIA_againstDifferentiallyPrivate,MIA_PrivacyRisks_of_SecuringML_againstAdversarialExamples,MIA_Liu2019SocInfMI}}&
\makecell[l]{\cite{MIA_Long2017TowardsMeasuringMembershipPrivacy,Overfitting_PrivacyRiskInML_AnalysingConnectionToOverfitting,Overfitting_UnintendedConsequencesOfOverfitting}\\ \cite{Reconstruction_PrivacyInPharmcogenetics, UnderstandingMI_onWellGeneralisedML}}& \makecell[l]{\cite{LOGAN_v1,LOGAN_v2,ExploitingUnintendedFeatureLeakage_inCollaborativeLearning} \\ \cite{GAN-Leaks, MIA_DeepModelsUnderGAN, ReID_usingGenerativeModels}}&\cite{Segmentations-Leak}& \cite{MIA_Salem2019MLLeaksMA,MIA_DemystifyingMIAsMLasService, MIA_ML, MIA_Liu2019SocInfMI}\\

Model Extraction Sec. \ref{sec:MEA}& \cite{ModelExtraction_CopycatCNN_withRandomNon-LabelledData,ModelExtraction_StealingMLmodelsViaAPIs,ModelExtraction_Jagielski2020HighAccuracyHighFidelityExtraction,ModelExtraction_Orekondy2019KnockoffNS,Reconstruction_Oh2018TowardsReverseEngineeringBlackBox,ModelExtraction_ModelReconstructionFromModelExplanation,ModelExtraction_ConnectionBetweenActiveLearningAndModelExtraction,ModelExtraction_Defence_PRADA_againstDNNstealingAttacks}& \cite{ModelExtraction_Krishna2020_modelExtractionOfBertBasedAPIs}& \cite{ModelExtraction_StealingHyperparameters,ModelExtraction_StealingMLmodelsViaAPIs, ModelExtraction_ConnectionBetweenActiveLearningAndModelExtraction}& - & \cite{ModelExtraction_CopycatCNN_withRandomNon-LabelledData,ModelExtraction_StealingHyperparameters,ModelExtraction_StealingMLmodelsViaAPIs,ModelExtraction_Jagielski2020HighAccuracyHighFidelityExtraction,ModelExtraction_Orekondy2019KnockoffNS,Reconstruction_Oh2018TowardsReverseEngineeringBlackBox,ModelExtraction_ModelReconstructionFromModelExplanation,ModelExtraction_ConnectionBetweenActiveLearningAndModelExtraction,ModelExtraction_Krishna2020_modelExtractionOfBertBasedAPIs,ModelExtraction_Defence_PRADA_againstDNNstealingAttacks}& \cite{ModelExtraction_StealingHyperparameters,ModelExtraction_StealingMLmodelsViaAPIs, ModelExtraction_ConnectionBetweenActiveLearningAndModelExtraction}& - & -& \cite{ModelExtraction_StealingHyperparameters, ModelExtraction_StealingMLmodelsViaAPIs, ModelExtraction_ConnectionBetweenActiveLearningAndModelExtraction}\\

Property Inference Sec. \ref{subsec:PIAs}& \cite{PropertyInferenceAttacks_Ganju2018PropertyInferenceAttacks,poisoning-backdoor-federated,poisoning-federated-adversarial-lens,poisoning-model-with-provable-convergence} & \cite{pia-from-poisoning}& \cite{PropertyInferenceAttacks_Hacking_extractDataFromClassifiers, poisoning-federated-adversarial-lens,poisoning-model-with-provable-convergence,pia-from-poisoning} & - & \cite{PropertyInferenceAttacks_Hacking_extractDataFromClassifiers,PropertyInferenceAttacks_Ganju2018PropertyInferenceAttacks,poisoning-backdoor-federated,poisoning-federated-adversarial-lens,poisoning-model-with-provable-convergence,pia-from-poisoning} & -& -& -& -\\

Reconstruction Sec. \ref{subsec:reconstruct}&\makecell[l]{ \cite{,ModelExtraction_StealingMLmodelsViaAPIs, MIA_ML, Leaks_Song2020OverlearningRS} \\ \cite{Reconstruction_Oh2018TowardsReverseEngineeringBlackBox, ModelExtraction_ModelReconstructionFromModelExplanation}\\ \cite{,Reconstruct_collaborativeInference,Reconstruct_collaborative_multitaskDiscriminator,Reconstruct_inverseShadowModel_beatsAmazonMLaaS,Reconstruct_SecretRevealer_GANpriorGuidingInversion,Reconstruct_GradientLeakage_NEURIPS2019_60a6c400, Reconstruction_Salem2020UpdatesLeakDS_onlineLearning}} & \cite{Reconstruct_GradientLeakage_NEURIPS2019_60a6c400} & \cite{Reconstruct_defence_purification, Overfitting_UnintendedConsequencesOfOverfitting, Reconstruction_PrivacyInPharmcogenetics, Leaks_Song2020OverlearningRS}& \cite{Reconstruction_PrivacyInPharmcogenetics, ModelExtraction_StealingMLmodelsViaAPIs} &\makecell[l]{ \cite{ Overfitting_UnintendedConsequencesOfOverfitting, MIA_ML, Leaks_Song2020OverlearningRS,  ModelExtraction_StealingMLmodelsViaAPIs} \\ \cite{Reconstruct_inverseShadowModel_beatsAmazonMLaaS,Reconstruct_SecretRevealer_GANpriorGuidingInversion,Reconstruct_GradientLeakage_NEURIPS2019_60a6c400,Reconstruction_Salem2020UpdatesLeakDS_onlineLearning, Reconstruct_collaborativeInference}\\\cite{Reconstruct_defence_purification,Reconstruct_Defence_Composite_basedOnNoise_titcombe2021practical, reconstruct_defence_noPeek,  Reconstruction_Oh2018TowardsReverseEngineeringBlackBox,ModelExtraction_ModelReconstructionFromModelExplanation}}&
\makecell[l]{\cite{Reconstruct_GradientLeakage_NEURIPS2019_60a6c400, Overfitting_UnintendedConsequencesOfOverfitting, Reconstruction_PrivacyInPharmcogenetics} \\ \cite{Leaks_Song2020OverlearningRS,ModelExtraction_StealingMLmodelsViaAPIs}}& - & -& \cite{Reconstruct_inverseShadowModel_beatsAmazonMLaaS, MIA_ML, ModelExtraction_StealingMLmodelsViaAPIs}\\

\hline

\end{tabular}
\vspace{-2mm}
\caption{\mcorr{Taxonomy of the data leakage research, summarized by the type of the data and tasks. Sec.~\ref{sec:data_types}, \ref{sec:tasks_types}, and~\ref{sec:leakage}.}}
\label{tab1}
\end{center}
\vspace{-0.7cm}

\end{table*}

\subbold{Personal data} are defined in the Article $4(1)$ of the GDPR, \cite{GDPR}, and, in loose terms, means data that directly or indirectly relates to an identified or identifiable natural person. Personal data may be collected routinely for legitimate ends. E.g., a National Health Service (NHS) patients have their personal data processed during routine health checks. 

\subbold{Sensitive data} are defined by the GDPR as the personal data revealing racial or ethnic origin, political opinions, religious or philosophical beliefs; trade-union membership; genetic data, biometric and health-related data; data concerning person’s sexual orientation. Such data require stronger safeguards for processing, storage, transfer, etc.

\subbold{Non-personal data} Any personal data fall under GDPR \cite{GDPR} protection, which implies a dichotomy -- everything outside the scope of personal are \textit{non-personal data}. Thus, non-personal data become of the utmost importance for any data-driven research, analysis, and commercial applications.

There are methods to separate personal from sensitive data. However, since researchers often use unconsented data it is common to adopt a cautious approach and assume that all of the data provided for research, even anonymised, falls under the special \textit{sensitive data} category, i.e., $f_j = f^{*}_j$ and $d_i = d^{*}_i$ for all $j$ and $i$.
The challenge is to mitigate against the risk of leaking any type of personal sensitive data that could be directly linked back to a real individual's identity (at the level of training sample $d_i$, such as patient record,
user information, etc). 
The real-world risk of linkage back to identity is complex, and depends on multiple factors -- the frequency of data points, the size of source datasets, and the availability of public data to support re-identification.








\subsection{Leakage for different data types}\label{sec:data_types}
Types of data leakage are largely data-specific (Table~\ref{tab1}).

\subbold{Data leakage in text data} Examples are individuals' names, dates of birth, full postcodes, full or partial addresses, telephone numbers, unique identity numbers, and job titles. In the context of training ML models on such data, one can imagine a predictive model, leaking specific sensitive data entries, features or full data records when deployed, \cite{TheSecretSharer_EvaluatingUnintendedMemorisation, MIA_ML}.

\subbold{Data leakage in images} Examples are individuals' faces or other identifying features, or embedded disclosive metadata (e.g. sensitive text on images). When training an ML model with sensitive image data, a generative model such as \cite{GANs_NIPS2014_5ca3e9b1}, trained on X-rays with hand-written notes on them or re-identifiable bone/denture implants, might occasionally reproduce an identifiable training image look-alike. \mj{This type of leakage could apply for other types of image text -- names on security badges, car license plates, etc.}

\subbold{Data leakage in tabular data} Examples of data leakage are similar to text data; however tabular datasets are constrained to predefined variables, so re-identification risks can be more accurately estimated according to statistical disclosure risks, based on features such as the data sensitivity, population size, zero-value entries, etc (see \cite{App_DaaS_NHSStatisticalDisclosure}). Whilst, for instance, re-identifying patients from rare combinations of diagnoses is possible, statistical disclosure control \cite{App_DaaS_NHSStatisticalDisclosure, App_DaaS_ONS_GuidanceOnIntruderTesting} is relied upon to make such a possibility distant.

\subsection{Leakage for different tasks}\label{sec:tasks_types}
Privacy violations and mitigation of such violations is not only data-specific, but also task- and model-specific. Please refer to Table~\ref{tab1}. Below is a detailed (but by no means exhaustive) overview of ML tasks and corresponding models researched for privacy-preservation and violation.

\subbold{Classification} is widely used for real world applications \cite{Classification_VisualSearchEbay, ClassificationClustering_FaceNet,Classification_SurveyMedicalImageAnalysis_LITJENS201760} and
is the best-researched task in terms of leakage and privacy attacks. A number of different kinds of attacks have been explored for image classification, on computer vision benchmarks like MNIST \cite{ModelExtraction_ConnectionBetweenActiveLearningAndModelExtraction, UnderstandingMI_onWellGeneralisedML, ModelExtraction_StealingMLmodelsViaAPIs, Reconstruction_Zhang2020TheSecretRevealer, MIA_Salem2019MLLeaksMA, MIA_ML, Overfitting_PrivacyRiskInML_AnalysingConnectionToOverfitting, Reconstruction_DeepLeakage, ModelExtraction_Jagielski2020HighAccuracyHighFidelityExtraction, Reconstruction_Oh2018TowardsReverseEngineeringBlackBox, PropertyInferenceAttacks_Ganju2018PropertyInferenceAttacks, Reconstruct_collaborativeInference, Monte-Carlo-MIA_onGenModels, MIA_DeepModelsUnderGAN, ModelExtraction_Defence_PRADA_againstDNNstealingAttacks, ModelExtraction_ModelReconstructionFromModelExplanation, ModelExtraction_Papernot2017PracticalBA, MIA_Rahman2018MembershipIA_againstDifferentiallyPrivate, MIA_DemystifyingMIAsMLasService, Reconstruction_Wang2019UserLevelPrivacyLeakageFromFederatedLearning}, CIFAR-100 \cite{DifferentialPrivacy_Jayaraman2019EvaluatingDP, MIA_Nasr2019ComprehensivePA, MIA_Salem2019MLLeaksMA, MIA_ML, Overfitting_PrivacyRiskInML_AnalysingConnectionToOverfitting, Reconstruction_DeepLeakage} and ImageNet \cite{ModelExtraction_Jagielski2020HighAccuracyHighFidelityExtraction, Reconstruction_Oh2018TowardsReverseEngineeringBlackBox, sablayrolles19a_BayesOptimalStrategiesforMIA}, as well as more applied datasets/tasks, such as classification of potential customer value \cite{MIA_Nasr2019ComprehensivePA, MIA_ML, MIA_DemystifyingMIAsMLasService}, classification of the income level based on the Census data \cite{ModelExtraction_ConnectionBetweenActiveLearningAndModelExtraction, PropertyInferenceAttacks_Ganju2018PropertyInferenceAttacks, UnderstandingMI_onWellGeneralisedML, ModelExtraction_StealingMLmodelsViaAPIs}, diagnosing breast cancer \cite{ModelExtraction_ConnectionBetweenActiveLearningAndModelExtraction, UnderstandingMI_onWellGeneralisedML, ModelExtraction_StealingMLmodelsViaAPIs} and classifying X-rays \cite{Reconstruction_Zhang2020TheSecretRevealer, Regression_PPOnmedicalImages}. There has been slightly less research on leakage from classifiers trained with tabular/mixed feature data 
\cite{ModelExtraction_ConnectionBetweenActiveLearningAndModelExtraction, Reconstruction_ModelInversionAttacks, Reconstruction_ModelInversionForPredictionSystems, ModelExtraction_StealingMLmodelsViaAPIs, ModelExtraction_StealingHyperparameters, ExploitingUnintendedFeatureLeakage_inCollaborativeLearning, MIA_Salem2019MLLeaksMA, MIA_ML,MIA_Nasr2019ComprehensivePA, PropertyInferenceAttacks_Hacking_extractDataFromClassifiers}, and even less involving time-series. A number of works have targeted UCI's diabetes dataset \cite{DiabetesDataset_UCI}, mainly for model extraction attacks \cite{ModelExtraction_ConnectionBetweenActiveLearningAndModelExtraction, ModelExtraction_StealingMLmodelsViaAPIs, ModelExtraction_StealingHyperparameters}, and also binary classification of text \cite{ExploitingUnintendedFeatureLeakage_inCollaborativeLearning}.

\subbold{Regression/Prediction} of unknown/future values of data samples has broad application in fields such as forecasting for financial and medical time-series. Leakage for regression was investigated for financial and medical time-series \cite{ModelExtraction_StealingMLmodelsViaAPIs, ModelExtraction_ConnectionBetweenActiveLearningAndModelExtraction, ModelExtraction_StealingHyperparameters}, numerical tabular data \cite{Overfitting_PrivacyRiskInML_AnalysingConnectionToOverfitting, ModelExtraction_StealingHyperparameters, Reconstruction_ModelInversionForPredictionSystems, ModelExtraction_StealingHyperparameters}, as well as mixed feature tabular data \cite{Reconstruction_PrivacyInPharmcogenetics, Overfitting_PrivacyRiskInML_AnalysingConnectionToOverfitting}.

\subbold{Generation/Synthesis} of realistic high quality data could solve the shortage of open-access data in the medical and financial domains. However, ensuring convincing privacy guarantees for generative methods is not a trivial problem \cite{Synthesis_medicalData_Tucker2020GeneratingHS, Defences_SyntheticData_PrivacyMirage_2020}. A good generative model should capture the underlying distribution of the real data \cite{GANs_NIPS2014_5ca3e9b1}, risking accidentally producing a doppelganger of a sensitive record (or a close enough sample). Simply sampling such a model could reveal individual records and attributes \cite{MemGANsWorkshop, MemorisationPrecedesGeneration}. A number of linkage attacks are developed for GANs \cite{LOGAN_v1, LOGAN_v2, GAN-Leaks, Monte-Carlo-MIA_onGenModels}, with some defences proposed for all data types \cite{DP_GAN, DP_GAN_forTimeSeries, DP_FunctionalMechanismForGANs, MemGANsWorkshop}. 





\subbold{Segmentation} is crucial for computer vision tasks. The privacy risks of sharing a medical image segmentation model publicly have been studied, e.g. by \cite{Segmentations-Leak}, for linkage attacks, showing that most state-of-the-art semantic segmentation models are vulnerable. Segmentation models' vulnerability with less than white-box access remains unexplored.

Privacy preservation has been very sparsely verified for any other tasks, but may also be important for tasks such as clustering, translation, transfer, and collaborative learning.



\subsection{How do user actions affect leakage?}\label{sec:user}

We differentiate between two types of users, defined below.

\subbold{Passive / honest-but-curious user} interacts with the trained model as intended by design and in compliance with protocols. All they can reveal is \textit{involuntary / benign leakage}, if the model has any such vulnerability.

\subbold{Malevolent user / an adversary} attempts to take advantage of potential vulnerabilities in the trained model, such as memorization and overfitting, aiming to extract sensitive data via \textit{privacy attacks}.

\vspace{-0.2cm}

\section{Involuntary Data Leakage}\label{sec:leakage}

Ways in which data leak without malicious user intervention include overfitting and memorization\cut{, and feature leakage}. \mcorr{Note that whilst overfitting implies some degree of memorization, memorization can occur while the model is still learning, i.e., before overfitting begins to happen \cite{TheSecretSharer_EvaluatingUnintendedMemorisation}. One or both of these can be the cause of data leakage.} 

\vspace{-0.2cm}
\subsection{Plain Overfitting} \label{sec:overfitting}

The hallmark of model overfitting is substantially higher accuracy on the training data than on the test data, usually caused by overtraining or unnecessarily large models being trained on smaller datasets \cite{Memorisation_Generalisation_ICLR2017}. 

\cut{\subbold{Risks}} The \mcorr{formalization of the} relationship between overfitting and privacy risks \cut{is not yet completely clear, due to the lack of}\mcorr{lacks} research on exactly how overfitting aids various data and model attacks
\cite{Overfitting_PrivacyRiskInML_AnalysingConnectionToOverfitting, Overfitting_UnintendedConsequencesOfOverfitting}. \mcorr{So far} it has been shown to be a sufficient but not strictly necessary condition for aiding membership inference and model inversion attacks (Sec.~\ref{subsec:MIAs} and~\ref{subsec:reconstruct}). 
\mj{Overfitting \cut{inevitably
results in a privacy loss for classification models}\mcorr{has also been shown to impact the privacy properties of classifiers}; \cite{Overfitting_UnintendedConsequencesOfOverfitting} formalizes the connection between the attacker model's inference advantage and the target model's generalization error for both membership inference and attribute inference attacks (Sec.~\ref{subsec:MIAs}).}

\cut{\subbold{Preventing overfitting}}For most models (but not all GANs \cite{GANs_NIPS2014_5ca3e9b1}) overfitting can be prevented simply monitoring the generalization error.

Nonetheless, overfitting is but one of the possible reasons for data leakage. Even stable, well-generalized models can leak sensitive data, \mj{e.g., due to memorization \cite{MIA_Nasr2019ComprehensivePA, UnderstandingMI_onWellGeneralisedML, TheSecretSharer_EvaluatingUnintendedMemorisation}. Specific model types and architectures, as well as the training dataset features also have an impact on leakage \cite{MIA_ML}.} 

\subsection{Memorization}\label{sec:mem}

Memorization of specific training data samples occurs when the model assigns some sample a significantly higher likelihood than expected by random chance \cite{TheSecretSharer_EvaluatingUnintendedMemorisation}. It raises serious privacy and legal concerns for sharing trained ML models publicly or providing them as a service \cite{Memorisation_AlgorithmsThatRemember_andLaw, Memorisation_MLmodelsThatRememberTooMuch}.
\cut{\subbold{Potential risks}} Potential risks include membership inference attacks, sensitive attribute and training dataset reconstruction 
\cite{Memorisation_MLmodelsThatRememberTooMuch}.

\cut{\subbold{Preventing memorization}} Although there is little research on preventing memorization, some evidence suggests that data augmentation reduces (but does not eliminate) the memorization capacity of a network, whereas increasing the size of the architecture increases its memorization capacity \cite{DEJAVU}. \mcorr{Specifically, \cite{DEJAVU} estimates the memorization in lower layers of convolutional neural networks, showing that 
fine-tuning of the upper layers might be insufficient to prevent memorization.} For GANs \cite{GANs_NIPS2014_5ca3e9b1, GANs_Improved_NIPS2016_8a3363ab}, \cite{MemGANsWorkshop} suggests that limiting the number of noise vectors at training time reduces memorization.





\subbold{Feature Leakage} \label{sec:featureLeak} \mcorr{Can be defined as a special case of memorization, which} occurs when sensitive attributes/features $f^{*}_j$ \mcorr{(rather than data samples $d_i$)} of the training data $D$ are unintentionally memorized and revealed by the trained model at inference time.



\cut{\subbold{Potential Risks.}} Feature leakage implicitly enables property inference attacks (Sec.~\ref{subsec:PIAs}). For instance, \cite{ExploitingUnintendedFeatureLeakage_inCollaborativeLearning} focuses primarily on leakage of \textit{unintended} features, i.e. inferring properties that hold for some subset of the training data but not in general for the entire class, which are also not necessarily the properties that the target model intended to capture in the first place. They show that property inference attacks are a danger for collaborative learning models (Sec.~\ref{subsec:PIAs} and~\ref{sec:defence_federated}). \mcorr{Feature leakage can be detected even when present in a few  \cite{yang2022understanding} or a single image \cite{hartley2022measuring}, with common techniques (e.g., augmentations, unlearning) shown to not offer protection.}

\cut{\subbold{Preventing feature leakage}} Interestingly, \cite{Leaks_Song2020OverlearningRS} discovers that \textit{overlearning}, i.e. the model learning attributes that are not part of the original objective or that make it sensitive to certain biases, can lead to feature/attribute leakage, and the model\cut{ capacity for being re-purposed for a privacy-violating agenda} vulnerability even in the absence of the original training data. Importantly, \cite{Leaks_Song2020OverlearningRS} also shows that overlearning cannot be prevented by merely censoring out the unnecessary attributes, meaning that certain defences, e.g., data obfuscation (Sec.~\ref{subsec:obfuscation}) will not reliably prevent overlearning. 


\begin{table*}[htbp]

\begin{center}
\begin{tabular}{l | c l | l c}
\hline
\textbf{Name} & \textbf{Target} & \textbf{Assumptions} & \textbf{Input}& \textbf{Result}\\
\hline
\hline
(a) Membership Inference, Sec.\ref{subsec:MIAs}&  &  & \makecell[l]{query dataset $D'$} & $d_i\in / \notin D$\\
\cline{4-5}
\makecell[r]{(M.I.: Attribute Inference)} & \makecell{training samples $d_i$\\ \\} & \makecell[l]{black-box access to $M_\theta$\\ \\ } &\makecell[l]{incomplete information \\ about a data point $d_i$} & missing details on $d_i$ \\
\hline
(b) Model Extraction, Sec.\ref{sec:MEA} (1) & \makecell{\\trained model} & \makecell[l]{model architecture \\ or type known} & \makecell[l]{labelled query dataset $D'$} & model parameters $\theta$\\
\cline{3-5}
\textcolor{white}{(b) Model Extraction,} Sec.\ref{sec:MEA} (2) & \makecell{$M_\theta$\\} & \makecell[l]{black-box access to $M_\theta$,\\ labelled query $D'$} & labelled query dataset $D'$ & model $M_\theta$ architecture \\
\cline{2-5}
\textcolor{white}{(b) Model Extraction,} Sec.\ref{sec:MEA} (3) & \makecell{functionality of the \\trained model $M_\theta$} & \makecell[l]{black-box access to $M_\theta$,\\ unlabelled dataset $D'$} & unlabelled query $D'$ & \makecell[l]{a model $M^{*}_{\theta^*}$, where \\ $M^{*}_{\theta^{*}}(x) \approx M_{\theta}(x)$} \\
\hline
(c) Property Inference, Sec.\ref{subsec:PIAs} & \makecell{whether $M_\theta$ exhibits\\ feature $f^{*}_i$} & white-box access to $M_\theta$ & weights $\theta$ of trained $M_\theta$ & $f^{*}_i\in / \notin {D, M_\theta}$ \\
\hline
(d) Reconstruction, Sec.\ref{subsec:reconstruct}& \makecell{reconstructing $D$\\(fully or partially)} & \makecell[l]{query access to $M_\theta$,\\ sometimes publicly \\ available $D'$ and $M'_\theta$} & \makecell[l]{publicly available or\\ generated query dataset $D'$} & \makecell[c]{training data $D$\\ (full or partial)}\\

\hline

\end{tabular}
\vspace{-2mm}
\label{tab2}
\end{center}

\end{table*}

\section{Malicious Leakage / Privacy Attacks}\label{sec:attacks}
To elaborate on Sec.~\ref{sec:user}, we define \textit{malicious leakage} (a term used interchangeably with \textit{privacy attacks} in this survey) as the actions of a malevolent user, an adversary who tries to take an advantage of trained ML model $M_{\theta}$, which we call the \textit{target model}, at inference time.
In this section we assume that the adversary has no access either to the original training data or to the training process of the target model. 
\mcorr{However,} the adversary's access to the trained model $M_{\theta}$ can be\cut{ either} \textit{black-box}, \textit{white-box}, or anywhere in between. \cut{ Note that s}Some\cut{ of the} methods \cut{reviewed }in this chapter also assume access to open-source data $D'$ that might or might not come from a similar distribution to the original (potentially sensitive)\cut{ training} data $D'$.

\begin{figure*}
\vspace{-0.2cm}
  \begin{subfigure}{0.22\textwidth}
    \includegraphics[width=\linewidth]{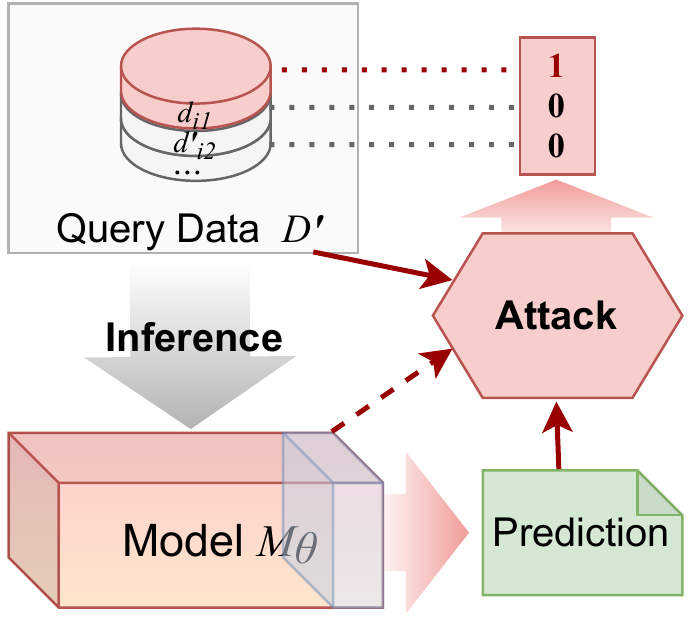}
    \caption{MIA} \label{fig:MIA}
  \end{subfigure}%
  \hspace*{\fill}   
  \begin{subfigure}{0.22\textwidth}
    \includegraphics[width=\linewidth]{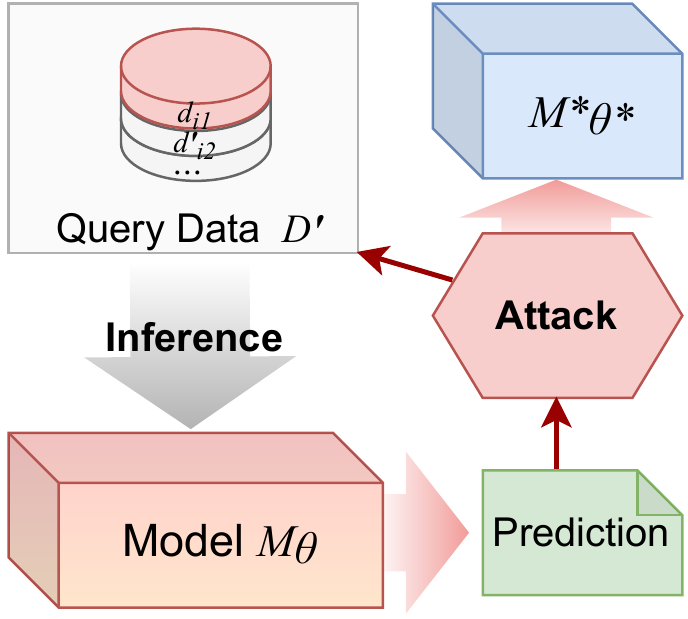}
    \caption{MEA} \label{fig:MEA}
  \end{subfigure}%
  \hspace*{\fill}   
  \begin{subfigure}{0.25\textwidth}
    \includegraphics[width=\linewidth]{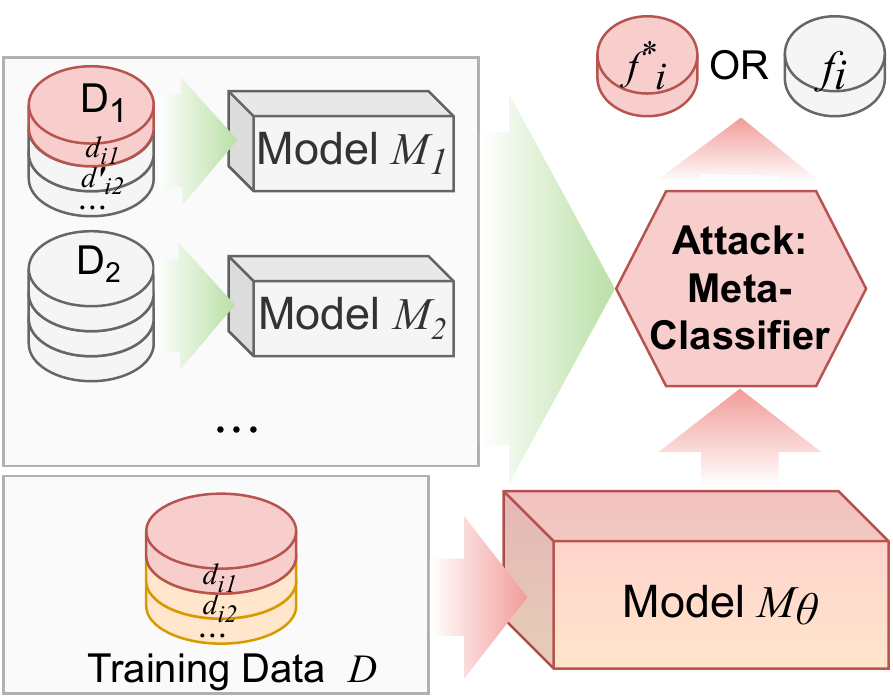}
    \caption{PIA} \label{fig:PIA}
  \end{subfigure}
  \hspace*{\fill}
    \begin{subfigure}{0.24\textwidth}
    \includegraphics[width=\linewidth]{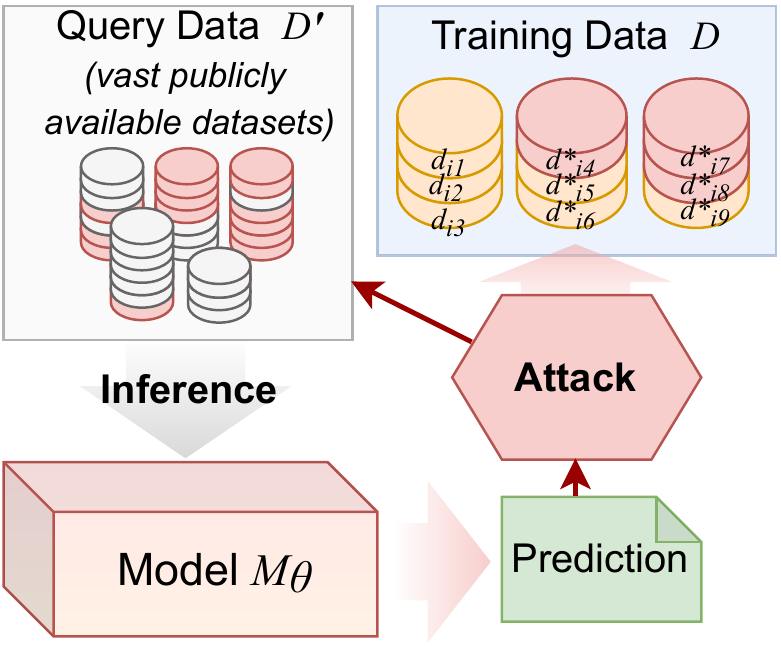}
    \caption{Reconstruction Attack} \label{fig:Reconstruction_attack}
  \end{subfigure}
\vspace{-0.2cm}

\caption{\mcorr{Most common types of the inference-time attacks:
\textbf{(a)} Membership inference attack (MIA), Sec.~\ref{subsec:MIAs}; 
\textbf{(b)} Model Extraction Attack (MEA), Sec.~\ref{sec:MEA};
\textbf{(c)} Property Inference Attack (PIA), Sec.~\ref{subsec:PIAs};
\textbf{(d)} Reconstruction Attack, Sec.~\ref{subsec:reconstruct}.}
} \label{fig:1}
\vspace{-0.5cm}
\end{figure*}

\subbold{Attacks Exploiting Memorization and Overfitting}
\cut{This is not an explicit class of privacy attacks; rather, m}Most attacks have a higher chance of success when overfitting comes into play\cut{, and several will implicitly exploit it} \cite{Memorisation_MIA_2020StolenMemoriesML, ExploitingOverfitting_Yeom2020OverfittingRA}. Overfitting alone has been shown to be enough for membership inference and more complex attribute inference attacks to succeed \cite{Overfitting_PrivacyRiskInML_AnalysingConnectionToOverfitting, ExploitingOverfitting_Yeom2020OverfittingRA} (see Sec.~\ref{subsec:MIAs}).

Other examples of exploiting memorization and overfitting apply to settings such as \textit{collaborative} (also known as \textit{federated}) \textit{learning} \cite{ExploitingUnintendedFeatureLeakage_inCollaborativeLearning}, where model \mj{gradient} updates can \mj{be used by the adversary -- the malicious participant -- to} leak sensitive information\mj{. Since the adversary provides part of the training data for the target model, the inference attacks (Sec.~\ref{subsec:MIAs}) are simplified to a supervised learning problem.
}

Another malicious setting, explored by \cite{Memorisation_MLmodelsThatRememberTooMuch}, features an adversary model provider (DaaS setting, Sec.~\ref{sec:daas}), supplying the model $M$ to a data owner, and receiving back a trained model $M_{\theta}$. Model architectures designed by \cite{Memorisation_MLmodelsThatRememberTooMuch}, could deliberately memorize the original training data, while maintaining reasonable performance on tasks like face recognition, image classification, and text analysis, even without the adversary directly controlling the training.

\subsection{Membership Inference Attacks}\label{subsec:MIAs}
ML models currently do not fall under GDPR protection. Nonetheless, advances in certain types of attack, such as \textit{membership inference attacks} (MIAs, also sometimes called \textit{``linkage attacks''}) and \textit{reconstruction MIAs}$^{1}$ 
can be used to identify the individual records used for training open-access ML models. Hence, MIAs can threaten user data privacy, supporting the argument that ML models should be classified according to their sensitive training data content \cite{Memorisation_AlgorithmsThatRemember_andLaw}.

\subbold{Formalization:} MIA (see Fig.~\ref{fig:MIA}) is a type of attack (lying anywhere in the range from white-box to black-box), that assumes the attacker has access to both:
\begin{itemize}
    \item \textit{The trained target model $M_{\theta}$} -- the more information about the model is available, the easier to attack. The adversary must at least have query access.
    \item \textit{Some query dataset $D'$} -- ideally containing the training data samples $d_i$, that have potentially been used for training $M_{\theta}$, i.e. $d_i \in D$ (as well as $d_i \in D'$). The adversary must at least have a dataset containing samples $d_i$ similar in distribution to those in the original $D$.
\end{itemize}

\noindent The target of MIA is to re-identify which of the samples $d_i$ were used for the training of the target model $M_{\theta}$. 

\mcorr{MIAs are usually performed either via \textit{shadow models}\footnote{\textit{Shadow model} is a term used in privacy attacks, in which a new model is trained by an adversary to mimic the behaviour of the target model, based on its query-output pairs.} \cite{MIA_ML, MIA_OnThePrivacyRisksOfModelExplanations, MIA_Salem2019MLLeaksMA, MIA_DemystifyingMIAsMLasService, MachineUnlearning_geopardisesThePrivacy} or based on comparison of \textit{metrics of the target model predictions} with an empirically chosen threshold. Such metrics can include prediction correctness\cite{Overfitting_PrivacyRiskInML_AnalysingConnectionToOverfitting, sablayrolles19a_BayesOptimalStrategiesforMIA, Label-onlyMIA, Metrics_QuantifyingMembershipInferenceVulnerability}, confidence level \cite{MIA_Salem2019MLLeaksMA}, entropy \cite{Metrics_SystematicEvaluationOfPrivacyRisksInML}, etc. For a detailed review and taxonomy of MIAs please refer to \cite{Survey_MIAs}.}

\subbold{Risks of query access to $M_{\theta}$.} While large companies take advantage of their user databases and deploy ML models on a large scale, there is always a risk of re-identification or misidentification of a user, given (even just query) access to the model. Offering ML as a service, i.e. providing the trained models in open and semi-open access, increases such risks. \cut{Vulnerability under MIAs is largely data-driven and hence data-specific. }Also, MIAs can be performed with just black-box access to the model \cite{sablayrolles19a_BayesOptimalStrategiesforMIA}, and without knowledge about the structure of the target model \cite{MIA_DemystifyingMIAsMLasService, MIA_Salem2019MLLeaksMA}. For example, \mcorr{a metric-based} \cite{MIA_Salem2019MLLeaksMA} shows that MIAs can succeed even without knowledge of the target model structure and without assuming that the query and original training datasets should come from the same or similar distributions, \mj{using just the posterior $M_{\theta}(d_i)$ of the target point $d_i$ and the empirically chosen threshold (based on attackers' priorities and query datasets available)}. \mcorr{Furthermore, \cite{Label-onlyMIA} introduces two label-only MIAs that require no confidence scores of target model $M_\theta$, instead directly assessing the robustness of the hard output labels under the input perturbations.}

\subbold{MIAs and Overfitting.} In addition to direct information about the model type, architecture, or parameter values (black- vs white-box MIAs), overfitting and poor generalization can significantly impact the vulnerability of a model. In fact, MIAs are \mj{likely} to succeed on an overfitted model even with only black-box access. \mj{For larger class-balanced multi-class datasets, \cite{MIA_ML} reports over 70\% attack accuracy for model overfit to a train-test accuracy gap of over 12\%, and up to 100\% attack accuracy for over 25\% gap.} Further, \cite{Overfitting_PrivacyRiskInML_AnalysingConnectionToOverfitting} and \cite{UnderstandingMI_onWellGeneralisedML} provide theoretical and empirical evidence that overfitting alone is sufficient to increase the attacker's success in performing MIAs. The same is proven by \cite{Overfitting_PrivacyRiskInML_AnalysingConnectionToOverfitting} for the \textit{attribute inference}\footnote{\textit{Attribute inference attack} (or \textit{reconstruction attack}) assumes access to the trained ML model and incomplete information about a data point, and aims to infer the missing information about that point \cite{Reconstruction_PrivacyInPharmcogenetics}.} attacks. However controlling overfitting (by minimizing generalization error) does not necessarily prevent a successful membership inference. Further, strategies for attacking well-generalised models via identifying the vulnerable target records and exploiting their influences on the target model are presented in \cite{UnderstandingMI_onWellGeneralisedML}. Finally, \cite{Overfitting_PrivacyRiskInML_AnalysingConnectionToOverfitting} shows that the possibility of attribute inference implies the possibility of membership inference, thereby making a connection between MIAs and reconstruction attacks (see Sec.~\ref{subsec:reconstruct}).

\subbold{Applications} of MIAs are numerous, both in terms of the types of the data and the models vulnerable to them. Applications which have been explored include: medical data \cite{MIA_Liu2019SocInfMI}, location data \cite{MIA_OnAggregateLocationData}, including time-series \cite{UnderTheHoodOfMIAonAggregateLocationTS}, translation systems \cite{MIA_hisamoto-etal-2020-membership_MachineTranslation}, collaborative learning (especially when the adversary performs as one of the participants) \cite{MIA_DeepModelsUnderGAN, MIA_DemystifyingMIAsMLasService}, and generative models for various types of data synthesis. The latter are usually GAN-based, where the discriminator of a \textit{shadow model} is often used for re-identifying the original training data samples in the query dataset \cite{LOGAN_v1, LOGAN_v2, GAN-Leaks}.

\subbold{Measuring} the success of MIAs is easy compared to other privacy attacks. The common metric is re-identification score -- the ratio of training and additional data samples in the query dataset, that have been correctly identified by MIA, or some modification of this metric \cite{MIA_Long2017TowardsMeasuringMembershipPrivacy, ReID_usingGenerativeModels}.

\subsection{Model Extraction Attacks}\label{sec:MEA}

\textit{Model Extraction Attacks (MEAs)} are not designed to steal the training data $D$ (although it is often a by-product of this class of attacks \cite{ModelExtraction_CopycatCNN_withRandomNon-LabelledData}); instead their end-goal is to steal the trained model functionality, see Fig.~\ref{fig:MEA}.

\subbold{Formalization of assumptions for different kinds of MEAs.} Model functionality can be captured in a few ways. From most to least prior knowledge required, MEAs can:
\begin{enumerate}[leftmargin=0.5cm]
    \item Steal the model parameters $\theta$, assuming the model architecture (or at least the type) is known to the attacker.
    \item Steal \textit{the entire model architecture $M_{\theta}$} when it is unknown -- a black-box-style model extraction attack.
    \item Steal the model functionality -- an extraction attack does not necessarily have to reverse engineer the target model itself. It might be enough to copy the functionality of it, e.g. \textit{make a different model $M^{*}_{\theta^*}$, where $M^{*}_{\theta^{*}}(x) \approx M_{\theta}(x)$}, where $x$ is some data plausible for a task domain at the inference time. This class of techniques can succeed without any assumptions on the model architecture or anything else, except query access to the target model.
    
\end{enumerate}


We now expand the above, ordering from greatest to least stringent requirements for attacker's prior knowledge.

\subbold{1) Stealing parameters $\theta$ and hyperparameters $\theta'$ of the ML models of the known class.} This setting assumes that an attacker is in possession of the most granular level of knowledge about the target model $M_{\theta}$ across all ME types.

For instance, \cite{ModelExtraction_StealingHyperparameters} assumes full white-box access to $M_{\theta}$, i.e. Machine Learning as a Service setting (MLaaS), where the adversary knows everything: the original training dataset $D$, the ML algorithm (an objective function) of the target model, and (optionally) the learned parameters of the target model $\theta$. Under these assumptions, \cite{ModelExtraction_StealingHyperparameters} proposes a method for efficiently stealing the hyperparameters $\theta'$ of the target models with both theoretical assessment and empirical evaluation on Amazon Machine Learning service.

A black-box attack on parameters $\theta$ is possible even without an access to the original training data $D$, assuming knowledge about the model class, the confidence values provided as an output of the target model, and/or the ability to query arbitrary partial inputs. Two efficient ways of stealing hyperparameters with aforementioned assumptions are introduced in \cite{ModelExtraction_StealingMLmodelsViaAPIs}. These attacks are also successful when the confidence values are omitted from the target model output, as a privacy precaution. The reported speeds of extraction of the $100\%$-equivalent of the trained models from publicly available services, Amazon ML and BigML, (for logistic regression and decision tree target models), is between just over a minute to just over half an hour \cite{ModelExtraction_StealingMLmodelsViaAPIs}.


\subbold{2) Reverse engineering black-box models or functionally equivalent model extraction.} Here the assumption of an adversary knowing the model architecture is relaxed, making the extraction attack much harder but not impossible \cite{ModelExtraction_Jagielski2020HighAccuracyHighFidelityExtraction, ModelExtraction_CopycatCNN_withRandomNon-LabelledData}. However, there is still an implicit assumption that the adversary has access to some suitable unlabelled data for querying the target model, not necessarily from the same domain as the original training data, but from a rich enough distribution to expose the full target model functionality.

An intuitive approach in this setting, based on creating an imposter dataset $D'$ and then training a functional equivalent $M^{*}_{\theta^{*}}$ of the target model $M_{\theta}$ on it, is offered by both \cite{ModelExtraction_CopycatCNN_withRandomNon-LabelledData} and \cite{ModelExtraction_Orekondy2019KnockoffNS}.  Both papers query the target model (black-box CNN) $M_{\theta}$ with some random unlabelled data $D'$, asking the target model itself to label the new dataset. This results in an imposter dataset $D'$, theoretically containing the knowledge of the target network $M_{\theta}$. The ``copycat'' network $M^{*}_{\theta^{*}}$ is then trained on the imposter dataset $D'$, and should be able to reproduce the behaviour of the target model $M_{\theta}$, i.e., $M^{*}_{\theta^{*}}(x) \approx M_{\theta}(x)$, where $x$ is some data plausible for a task domain. The empirical results of \cite{ModelExtraction_CopycatCNN_withRandomNon-LabelledData} (for CNN class models) show at least $93.7\%$ attack accuracy on a variety of problems (measured as the ability perform in the same way as the target model), and $97.3\%$ of the performance when applied to the Microsoft Azure Emotion API. \cite{ModelExtraction_Orekondy2019KnockoffNS} shows between $92\%$ and $105\%$ performance of the target model. They explain the additional improvement on the target model by the regularizing effect of training on soft-labels, introduced as the ``soft targets'' in \cite{MEA_DistillingKnowledgeInNN}.

\subbold{3) Stealing functionality with minimal assumptions.} The next assumption to relax is access to the unlabeled query data. \cite{Reconstruction_Oh2018TowardsReverseEngineeringBlackBox} assumes no prior data knowledge and no knowledge of the class of $M_{\theta}$. Instead, they train a meta-model capable of inferring the architecture of $M_{\theta}$ and training hyperparameters (e.g. the optimization algorithm and training dataset) from a series of queries, hence turning the black-box target models into white-box models, making the target models susceptible to all of the above mentioned attacks.

Last but not least \mj{amongst the minimal assumption methods}, \cite{ModelExtraction_Jagielski2020HighAccuracyHighFidelityExtraction} explores the trade-off between accuracy and fidelity of MEAs, where accuracy stands for performing well on the underlying task, and fidelity for matching the target model predictions. They focus on high-fidelity, and claim the first practical functionally-equivalent model extraction, i.e. $M^{*}_{\theta^{*}}(x) = M_{\theta}(x)$, as well as faster querying, compared to competitors. \mj{This is achieved by a learning-based attack method, that utilizes the target model as an oracle for training the adversary model.}




\subbold{Model extraction for some more specific applications.} An important limitation of all of the aforementioned MEA-related research is that it focuses primarily on classification and prediction tasks. However, there are other interesting applications, e.g. \cite{ModelExtraction_ModelReconstructionFromModelExplanation} investigates model extraction attacks in a setting where the target model provides not only traditional outputs, but the gradients with respect to the input data as an explanation for its outputs. Active learning for model extraction in MLaaS settings is covered by \cite{ModelExtraction_ConnectionBetweenActiveLearningAndModelExtraction}, both for implementing model extraction attacks and investigating possible defences. They find that active learning is very similar to MEAs. There is also some exciting research on model extraction of natural language models, such as BERT -- \cite{ModelExtraction_Krishna2020_modelExtractionOfBertBasedAPIs} finds that not only simple query access to the target model is sufficient, but also that no real or semantically plausible data is required for querying the target model. Random sequence querying paired with a task-specific heuristic is enough for extracting approximate models for natural language inference and question answering.

Model extraction for generative models remains unexplored. One can argue that a principle similar to \cite{ModelExtraction_Orekondy2019KnockoffNS} and \cite{ModelExtraction_CopycatCNN_withRandomNon-LabelledData} could work, i.e. sampling the target model for random inputs (for instance conditions for the generator in GANs) in order to create a fake dataset for training a functionally identical model. However, to our knowledge there is no published work confirming this in practice.

\subsection{Property Inference Attacks} \label{subsec:PIAs}


\textit{Property Inference Attacks (PIAs)} constitute a type of attacks where an attacker tries to extract a specific sensitive attribute or feature of interest $f^{*}_i$ from a given target model $M_{\theta}$. See Fig. \ref{fig:PIA} for the overall structure.

\subbold{Assumptions and Formalization.} PIAs are white-box attribute inference attacks, assuming complete access to $M_{\theta}$ and its training information. They are based on the principle that similar models, trained on similar datasets, exhibit similar properties.
The goal is to build a meta-classifier $MC$, capable of telling whether a model $M_i$ contains an attribute $f^{*}_i$.
In order to train $MC$, an attacker trains a series of shadow classifiers, $M = \{M_1, .., M_n\}$ on some dataset, $D = \{D_1, .., D_n\}$, where only some of the subsets $D_i$ exhibit the property $f^{*}_i$.
The shadow models are not explicitly trained to learn the property $f^{*}_i$, but learn it as a consequence of the bias introduced in the dataset.
At inference time the $M_\theta$, trained on the original dataset $D_x$, is classified by $MC$ as either exhibiting $f^{*}_i$ or not.
Weights and biases of $M_\theta$ are often used as features in $MC$ training.

The first PIA was conducted on Hidden Markov Models and Support Vector Machines \cite{PropertyInferenceAttacks_Hacking_extractDataFromClassifiers}. Weights of the hidden states served as inputs for the HMMs while weights and biases of support vectors were used to train the meta-classifier for SVMs. The logical transition of this approach to fully connected networks was shown in \cite{PropertyInferenceAttacks_Ganju2018PropertyInferenceAttacks}, where the weights and biases of the neural networks were used as input to the meta-classifier. To account for the permutation invariance in the representations
, the meta-classifier itself was a network that learnt to account for all permutations of a particular neural network layer's weights and biases. This architecture was inspired by Deep Sets \cite{deepsets}.

\subbold{Poisoning Attacks} is a special case of PIAs, where an adversary can pollute the data $D$ or the model $M$ during the training, causing a bias in the $M_\theta$ output, resulting in a leakage. (\textit{Training time} attacks are out of scope of this work.)

\subbold{Applications} of PIAs are so far somewhat limited to fully connected neural networks. Moreover, no publications show PIAs applied to anything but classification tasks.

\subsection{Reconstruction / Model Inversion Attacks}\label{subsec:reconstruct}



\textit{Reconstruction / Model Inversion Attacks} are methods for partial reconstruction of private datasets from aggregated publicly available information $D'$, including open-access or query-only trained ML models $M_\theta$. See Figure~\ref{fig:Reconstruction_attack}.

\subbold{Applications.} 
Reconstruction attacks have been applied to a variety of scenarios, for instance, to the \textit{federated learning setting}, \cite{Reconstruct_collaborativeInference}, including an interesting application of GANs trained with a multitask discriminator that outputs the reality indicator for the data, its class, and user identity, \cite{Reconstruct_collaborative_multitaskDiscriminator}.

A variety of applications of model inversion exist in the \textit{general (centralized) setting}. E.g., \cite{Reconstruct_inverseShadowModel_beatsAmazonMLaaS} trains a second neural network as an inverse of the target model to perform the inversion, with its performance validated on Amazon Rekognition
.
Another GAN-based method, called \textit{generative model-inversion attack} \cite{Reconstruct_SecretRevealer_GANpriorGuidingInversion} uses GANs to learn the distributional prior of the data, which later guides the inversion process.
Finally, \cite{Reconstruct_GradientLeakage_NEURIPS2019_60a6c400} explores the \textit{Deep Leakage from Gradient}, an incredibly efficient inversion attack, accurate to the pixel-level for images and token-level for natural language, proving gradients of the model are unsafe to share publicly.

\textit{Reconstruction attacks in an online learning setting} have been studied in \cite{Reconstruction_Salem2020UpdatesLeakDS_onlineLearning}, where the adversary probes a model with a particular data point (\textit{MIA}, Sec.~\ref{subsec:MIAs}) or a particular set of data points (\textit{Group MIA}) before and after training a model with additional data, to assess how the model's outcome changes as a result of online training. The attacks follow a general encoder-decoder structure. 

\subbold{Defences.} Several defences have been proposed against reconstruction attacks: \cite{Reconstruct_noiseInterference} suggested the ``noise interference'' technique, which can render an invertible model non-invertible by adding noise. Another noise-based defence, this time for the federated learning setting, has been recently proposed by \cite{Reconstruct_Defence_Composite_basedOnNoise_titcombe2021practical}. They use a simple additive noise method and, interestingly, they find that pairing it with another existing method NoPeekNN, \cite{reconstruct_defence_noPeek}, improves the defence. For classifier target models, \cite{Reconstruct_defence_purification} suggests ``purifying'' the confidence score vectors of the target model by reducing their dispersion. This can help, since some of the MIAs and some of the reconstruction attacks use the target model confidence score vectors for guidance.

\begin{figure*}
    \centering
    \includegraphics[scale=0.82]{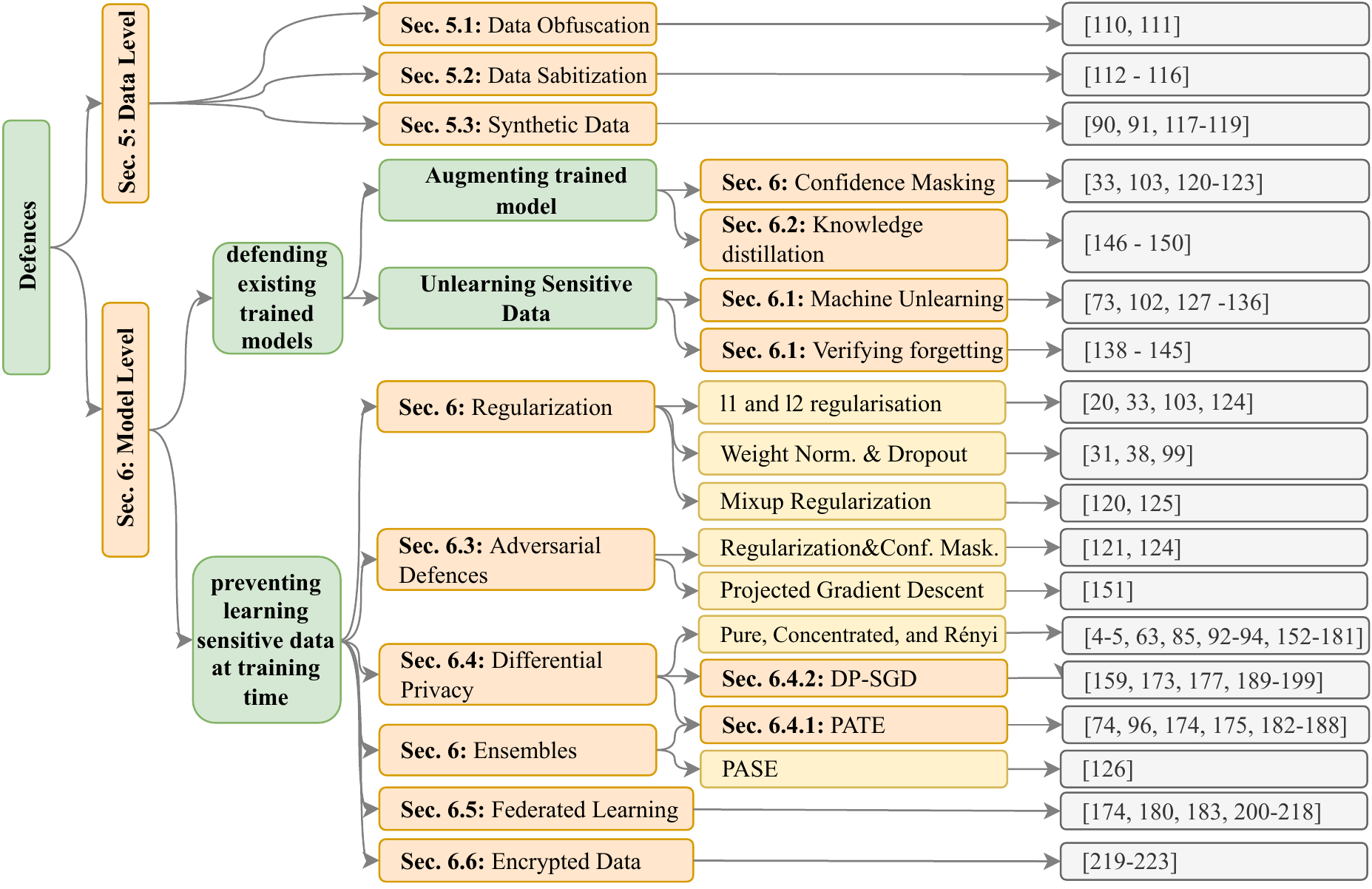}
    \vspace{-0.2cm}
    \caption{\mcorr{Taxonomy of the defence sections (Sections~\ref{sec:defences_data} and \ref{sec:defences_model}).}}
    \vspace{-0.5cm}
    \label{fig:DataLeaks_defences}
\end{figure*}

\section{Current Defences at Data Level}\label{sec:defences_data}
Defence methods aim to prevent malicious privacy attacks 
from succeeding. There are a few stages in model training and deployment where defences can be implemented. They can largely be dichotomized as applying augmentations at the \textit{training data level} versus training, tuning, and designing models with inbuilt defence mechanisms -- at the \textit{model level}. Refer to Figure~\ref{fig:DataLeaks_defences} for the proposed taxonomy of defences.

At data level, simply deleting sensitive features / entries can violate the data integrity and consistency and represent a privacy risk of its own, since the pattern of ``missingness'' might reveal some data properties. Hence data obfuscation and sanitization are often applied to mask, scramble, or overwrite the sensitive information with a realistic fake.

\subsection{Data Obfuscation}\label{subsec:obfuscation}
Data obfuscation perturbs the sensitive information in the data through either \textit{scrambling or masking} of some sort.

For instance, \cite{DefenceDataLevel_Zhang2018PrivacypreservingML} introduces an obfuscation function 
that addresses the trade-off between user privacy and service quality, which is dependent on the severity of the data perturbation. The adaptive mechanism anticipates and protects against optimal inference algorithms by designing a game between the obfuscation designer and the potential inference attack.
Meanwhile, \cite{DefencesDataLevel_Wang2020MIASecED} is concerned with the difference between a trained model's predictions on training and test data and the inference risks this difference presents. They suggest mitigating those risks by narrowing the dynamic ranges of the sensitive features in the training data, such that the training, test, and synthetic data are forced to have similar predictions by the same model.

\subsection{Data Sanitization}
Data Sanitization aims to disguise the sensitive information within the data by overwriting it with realistic-looking synthetic data, using techniques like flipping labels or adding noise of certain specifications. Recent developments also include randomization algorithms satisfying the $\epsilon$-differential data privacy criteria \cite{Sanitization_ImprovedSanitization_Zaman2017AnID}. Data sanitization is often a natural precaution for \textit{adversarial attacks} \cite{Sanitization_AgainstAdversarialAttacks_againstPoisoning_Chan2018DataSA} (a large class of training time attacks, which is out of the scope of this paper).

The aforementioned data modifications are limited by the assumptions made about data complexity.

Sanitization is a potential defence against the inference attacks on the social media networks, e.g., \cite{Sanitization_PreventingInferenceAttacks_socialMedia_Cai2018CollectiveDF} utilizes a collective manipulation sanitization techniques on the user profile and friend connection data to prevent inference attacks from identifying users from their friend connections. Additionally, \cite{Sanitization_forSocialNets_Tambe2016DataSF} argues that nouns convey most of the information in a sentence, hence sanitization can be conducted by treating nouns in the sensitive sentences as keywords that need overwriting with 
random entries. Sanitization applications include self-destruct data-processing cycles \cite{Sanitization_AutomaticallySelfDestructingData_Zhu2019FlashGhostDS}. These use threshold cryptography to overwrite data enough times to render it non-recoverable and hence ensure user data self-erase after a certain validity period.


\subsection{At Data Level: Learning with Synthetic Data}
Learning with synthetic data can be viewed as a natural extension to both data obfuscation and sanitization, since it involves perturbing/disguising the sensitive information. High-fidelity synthetic data, generated with privacy guarantees, could solve training data shortage problems for a wide variety of applications. 
It would allow open access to realistic synthetic data for researchers and facilitate internal data transfers within the organizations, where clients data cannot be shared across branches, divisions, hospitals, etc. 

There is evidence suggesting some already successful applications - generating high quality synthetic patient 
data \cite{Synthesis_medicalData_Tucker2020GeneratingHS} for testing ML healthcare software, using a combination of techniques including probabilistic graphical modelling. Another potentially useful approach is data synthesis via a differentially private autoencoder with empirical assessment of both the utility and quality of the results \cite{Defences_DiffPrivate_SyntheticData_Abay2019}.

There are several GAN-based models that are designed to produce synthetic data with certain privacy guarantees. For example, \cite{time-GAN} is meant for generation of time-series data with DP guarantees, and \cite{PrivGAN} is designed to preserve privacy under MIAs at a small performance trade-off. 

Although synthetic data might seem appealing as a remedy for the sensitive data leakage problem, it is not the case in reality, largely because a good generative model captures the underlying training data distribution, and might leak at the very least some of the properties of the dataset into its generated data, enabling PIAs. Moreover, \cite{Defences_SyntheticData_PrivacyMirage_2020} finds that generative models tend to store richer information, enabling attribute inference. Furthermore, they show that generative models are vulnerable to MIAs
, even when trained under differential privacy guarantees, perhaps because memorization in models like GANs and VAEs cannot be fully eradicated. Thus far synthetic data generation with privacy guarantees remains elusive. 

    

\section{Current Defences at Model Level}\label{sec:defences_model}

\mcorr{Model defence techniques can be loosely separated into those protecting an existing trained model from leaking learnt sensitive information vs those preventing models from learning such information in the first place (Fig.~\ref{fig:DataLeaks_defences}).}

\mcorr{As many of the inference time attacks rely on the confidence scores of the target model, defence can be simply perturbing these via confidence masking or regularization.}

\subbold{Confidence Masking}\mcorr{usually comprises one of: hiding confidence scores of the model output and providing only the final label \cite{Label-onlyMIA}, or showing only top-K confidence \cite{MIA_ML, MIA_andDefences_inClassificationModels}, perturbing the confidence scores directly \cite{jia2019memguard}. Finally, prediction purification \cite{Defence_Yang2020DefendingMI, Defence_MLCapsule_MLAS} used against inversion and membership inference attacks replaces the confidence scores with reconstructed privacy-preserving representations.}

\subbold{Regularisation}\mcorr{is a standard way to prevent overfitting and hence can be considered a defence against data leakage \cite{MIA_ML}. Standard types of regularization include l1 and l2 regularization \cite{MIA_ML, UnderstandingMI_onWellGeneralisedML, AdversarialDefences_AdversarialRegularisation, Label-onlyMIA}, weight normalization and dropout \cite{LOGAN_v1, MIA_Salem2019MLLeaksMA, Memorisation_MIA_2020StolenMemoriesML}, and mixup regularization \cite{ MIA_andDefences_inClassificationModels, Defence_MIA_Yin2021_againstMoreKnowledgeable_MIA}.}

\subbold{Ensembles}\mcorr{have also been used against privacy attacks, for example switching ensembles (PASE) have been shown to work against MIAs \cite{Defences_PASE}, and teaching ensembles (Sec. \ref{subsec:PATE}) work against a broad spectrum of attacks.}

The rest of this section
covers more complex defences.

\subsection{Machine Unlearning / Forgetting}


The General Data Protection Regulation (GDPR), \cite{GDPR}, enforced by the European Union in May 2018, is aimed at protecting user privacy. Amongst other things, GDPR ensures the user's right ``for the explanation'' about how their data are being stored and used, as well as the right ``to be forgotten'', i.e. a user can request their data to be deleted from a database. The natural next question: \emph{What if these data also had to be ``forgotten'' by the AI models powering a service?} 

The obvious course of action would be to remove the user data that needs to be forgotten from the training dataset and retrain the model from scratch. However, often the computational costs involved would make this an infeasible solution, creating a demand for techniques to unlearn the requested data and its traces from the trained models.



\subbold{Machine Unlearning.} This term was introduced by \cite{MachineUnlearning_formalisation} who proposed the need for a ``forgetting system'', and introduced one of the first unlearning algorithms based on converting learning algorithms into \textit{\mj{summation form}}\footnote{\mj{\textit{The summation form} is a technique where model weights are not trained on each data sample, instead they are trained on a small number of sums of the data sample transforms. Aforementioned transforms are achieved through pre-defined efficiently computable transformation functions. When the data sample is erased, these sums get re-computed, and the model is efficiently updated.}} for efficiently forgetting data traces. This method also works against data pollution attacks. The first framework for instantaneous data summarization with machine unlearning using a resilient streaming algorithm, involving submodular optimization was presented in \cite{MachineUnlearning_theRightToBeForgotten}; it comes with a constant factor approximation guarantee to the optimum solution. Further, \cite{MachineUnlearning_MakingAIForgetyou} provides formalization for machine unlearning in a variety of instances, and proposes an efficient unlearning algorithm for k-means clustering, with accompanying statistical analysis of the results. An unlearning algorithm for linear regression methods, based on the projective residual update and use of synthetic data points, was introduced by \cite{MachineUnlearning_ApproxDataDeletionFromML}. Later,
\cite{MachineUnlearning} proposed to limit the effect that a single data sample can have on the training process by training multiple models on subsets of the training dataset. This would imply storage and computational costs for retraining multiple models. 
In a similar attempt to limit the effect of a single data point on training \cite{MachineUnlearning_UnderstandingBlackBox} and \cite{ MachineUnlearning_ASwissArmy} suggest a Newton-based estimation of the effect of such data point on the model predictions. This estimate can be immediately used for guiding the machine unlearning.

A comparatively computationally light method
\cite{MachineUnlearning_LinearFiltrationForLogitClassifiers} suggests forgetting logit-based classifiers through linear transformation to the output logits. This would leave a data sample trace in the weights of a neural network model.
\cite{CertifiedDataRemovalFromML} focuses on data removal from differential privacy perspective, and provides an algorithm for convex problems, based on a second order Newton update, layered over a DP DNN.

An algorithm proposed by \cite{Golatkar2019EternalSO} conducts unlearning for DNNs trained with SGD, and is based on shifting the weight space of the model by adding noise to the weights. Specifically, \cite{Golatkar2019EternalSO} focuses on selective forgetting by ``scrubbing'' the weights of the neural net, so that it need not be trained from scratch, without requiring the access to data to be forgotten. Further, \cite{ForgettingOutsideTheBox} proposes weight scrubbing based on the Neural Tangent Kernel at the level of the model activations, which allows not only better handling of the null-spaces in network weights (which is essential for over-parameterised models like DNNs), but also for the ``one-shot'' forgetting to work better than \cite{Golatkar2019EternalSO}. This work introduces a new set of bounds that quantifies the average information per query an attacker can extract from a model.

\subbold{Verifying forgetting.}
There is a difference between deliberately unlearning the traces of information from a model versus verifying it has indeed been forgotten (intentionally or otherwise). Additional considerations: \textit{1)} Forgetting can occasionally happen on its own (\textit{``catastrophic forgetting''} in reinforcement learning); \textit{2)} different data samples bear varying amounts of unique information and contribute to the model final weights differently \cite{VerifyingForgetting_EstimatingInformativeness}; \textit{3)} forgetting a specific data entry (a single person's entry) in the training set and consequently its trace in the system is non-trivial, because of the possible \textit{trace overlap}.
Trace overlap means the updates (the trace) extracted from the entry to-be-forgotten is equivalent to the trace obtained from another record that is still a legitimate training sample, hence this trace should be kept in the model. In light of GDPR \cite{GDPR} and the ``right to be forgotten'' there is a lot of focus is on formalization and good ways of performing verification of forgetting \cite{FormalisingDataDeletion}. 

These intricacies have led to several directions of verifying forgetting. For instance,  
\cite{LiuHaveYouForgotten} focuses primarily on applying statistical methods, i.e. Kolmogorov-Smirnov distance, to find a discrepancy in the output distributions between a model that has supposedly ``forgotten'' certain traces and a reference shadow model (See Sec.~\ref{subsec:MIAs} about \textit{shadow models}), trained on different datasets to model forgetting with and without a \textit{trace overlap}.

In case if the ``core'' dataset, that should not be forgotten, is known, \cite{VerifyingForgetting_MixedPrivacy} offers an effective method of forgetting the traces of the additional data, that involves replacing a standard deep network with a suitable linear approximation.

There are also plenty of context-specific applications, including forgetting data for neural network predictors \cite{Forgetting_obliviousNNsPredictors} \mj{by applying carefully engineered oblivious protocols for commonly used operations on trained networks. F}or network embeddings \cite{VerifyingForgetting_NetworkEmbeddings} \mj{investigates forgetting a single node by removing the representation vector from the network embedding, and finds that often this is not sufficient since the information can be still encoded in the embedding
vectors of the remaining nodes.} For text generation models \cite{Forgetting_DataAuditing_forTextGenerationModels} \mj{suggests black-box model-auditing technique successful on well-generalized models (not overfitted to their training data), and} \cite{Forgetting_DataAuditing_blackBoxModels_withModelDistill} \mj{proposes a model-auditing method based on the model distillation and model comparison techniques}.

\subbold{Limitations and Risks.} Despite the benevolent intentions of machine unlearning, it should be applied with caution due to the risks involved. Analyzing the risks of data leakage (under MIA) for black-box classifiers after machine unlearning, \cite{MachineUnlearning_geopardisesThePrivacy} finds that in some cases the \textit{unlearnt} model can leak information about the forgotten data, even when the original \textit{non-unlearnt} model did not leak such information.

\subsection{Knowledge Distillation}

Knowledge distillation has been actively used to compress models and thus facilitate deployment on resource constrained devices, however, it can also be applied to preserve privacy. Distillation for Membership Privacy uses distillation to train models with membership privacy by leveraging various sources of noise in the model distillation process \cite{Defences_knowledgeDistillation}. Distillation-based methods based on the fast gradient sign method \cite{defenses_distillation-1st} and the Jacobian attack \cite{defenses-distillation-effectiveness} have been shown to train privacy preserving models where large perturbations to the input are required to make a distilled model cause a wrong prediction. However, \cite{defenses-distillation-failure-2} showed that distillation fails to mitigate attacks proposed in \cite{defenses-distillation-failure}.

\subsection{Adversarial Defences}\label{subsec:defences_adversarial}
Adversarial defence strategies use potential attack models as a penalty when training the target model $M_{\theta}$. Although in theory most privacy attacks can be used in some way during the training of $M_\theta$ as adversaries to defend against, in practice this setting has been mostly explored for MIAs.

\subbold{Adversarial Defences for MIAs.} A lot of research conducted on protecting against black-box MIAs with adversarial examples.
For example, \cite{AdversarialDefences_AdversarialRegularisation} anticipates a MIA, and regularizes the target model during the training via min-max game-based adversarial regularization, so that predictions of the target model on its training data are indistinguishable from its predictions on other data points from the same distribution. This technique not only claims membership privacy, but also -- good target model generalization. Memguard, \cite{jia2019memguard}, has been the first defence with formal utility-loss guarantees against black-box MIAs. 
It proposed adding carefully designed noise to the target model confidence score vectors, turning these into adversarial examples, that would make MIA classifier vulnerable.


\subbold{Limitations and Risks.} Interestingly, some of the proposed adversarial defences, such as projective gradient descent (PGD) adversarial training \cite{Leakage_Madry2018TowardsDL}, on the contrary increase the model’s susceptibility to membership inference attacks.
Theoretically, many of the privacy attacks could be used as adversaries to improve against during training of $M_\theta$. Nonetheless, one has to be careful, as \cite{MIA_PrivacyRisks_of_SecuringML_againstAdversarialExamples} has proven that using some state-of-the-art attacks as adversaries during training can weaken the defence against these and the new attacks compared even to the original undefended model.





\subsection{Training with Differential Privacy} \label{subsec:DP}
The idea behind \textit{differential privacy} (DP) is to gather confidential user data for analysis without compromising the confidentiality of each individual user. It was formally defined in 2006 by \cite{DifferentialPrivacy_AlgorithmicFoundation} -- the algorithm $K$ is considered to be $\epsilon$-private if for all datasets $D_1$ and $D_2$ differing in at most one data entry and all events $S$
\[ Pr[K(D_1)\in S] \leq exp(\epsilon) + Pr[K(D_2)\in S].
\]
This can be interpreted as follows: a differentially private algorithm's functionality should remain unchanged whether any single entry is or is not present in its training dataset. Thus, unlike some other defences, DP provides a guarantee on the maximum privacy loss: the maximum divergence between these two distributions (or a maximum log odds ratio for any event $S$) is bounded by the ``privacy budget'' $\epsilon$. 
This guarantee is known as ``pure'' differential privacy.

\subbold{Concentrated DP and R\'enyi DP.} There exist generalizations and relaxations of DP methods, that tend to enjoy higher accuracy than ``pure'' DP. For instance, $(\epsilon, \delta)$-differential privacy, \cite{DP_EpsilonDeltaRelaxation}, guarantees that with probability of at most $(1-\delta)$ the privacy loss does not exceed $\epsilon$. Typically this helps with the trade-off between privacy and accuracy of the model, and ``pure'' DP can be viewed as a special case when $\delta=0$. However, in the case of multiple queries, the bound grows, which is why \cite{DP_Dwork_ConcentratedDP} proposed \textit{Concentrated Differential Privacy (CDP)} relaxation, not only improving on the accuracy but also offering tighter bounds on the expected privacy loss for \textit{group privacy}. The privacy loss accounting, training efficiency and model quality can be improved using two different data batching techniques proposed by \cite{DifferentialPrivacy_ModelPublishing} as an extension to classic CDP. Further quantitative results for CDP were provided in \cite{DifferentialPrivacy_ConcentratedDP} by re-defining the concept of DP in terms of the R\'eényi divergence between the distributions obtained by running an algorithm on neighboring input, and defining \textit{zero-Concentrated Differential Privacy (zCDP)} with its corresponding lower bounds. An alternative approach is to adopt \textit{R\'enyi Differential Privacy (RDP)} proposed by \cite{DifferentialPrivacy_Renyi}, which claims more accurate analysis of the privacy loss due to another relaxation -- CDP requires a linear bound on all positive moments of a privacy loss variable, whereas \cite{DifferentialPrivacy_Renyi} definition applies to one moment at a time. Further, \cite{Privacy-Preserving_DP_subsampling} proves a tight upper bound on RDP for subsampling in DP, it also generalizes the results of the \textit{moments accounting technique} \cite{DifferentialPrivacy_DeepLearningWithDP}, to any RDP algorithm. The \textit{moments accounting technique} \cite{DifferentialPrivacy_DeepLearningWithDP} is a DP framework for deep learning that improved training computational efficiency by introducing algorithms for efficient gradient computation for individual training examples, sharding tasks into smaller batches to reduce memory footprint, and applying DP principal projection at the input layer. Tool-wise, this framework is built in Tensorflow \cite{Tensorflow}.

    
    

\subbold{Differential Privacy Surveys.} In addition to some of the aforementioned previous privacy reviews, \cite{Survey_SOK, Survey_OverviewOnPrivacy_ArXiv}, there exist several surveys focusing specifically on DP, from early works such as \cite{DifferentialPrivacy_Survey2008}, to \cite{DP_Survey_Ji2014DifferentialPA, DP_and_Federated_Survey_medical_Kaissis2020SecurePA}. We refer readers to these specialised surveys for a more complete picture of the field.

\subbold{Applications of DP with respect to different tasks.} The more traditional applications of DP, outlined in \cite{DifferentialPrivacy_AlgorithmicFoundation} are the DP online learning, \cite{DP_OnlineLearning_Jain2012DifferentiallyPO, DP_OnlineLearning_Thakurta2013NearlyOA, DP_OnlineLearning_ICML_AgarwalS17} and DP empirical risk minimization, \cite{DP_EmpRiskMinimisation, DP_EmpRiskMinimisation_fasterAndMoreGeneral, DP_EmpRiskMinimisation_ConfidenceIntervals, DP_EmpRiskMinimisation_withNonConvexLossFunc, DP_EmpRiskMinimisation_withNonConvexLossFunc_nonStationary, DP_EmpRiskMinimisation_withSparcityInducingNorms, DifferentialPrivacy_EmpiricalRiskMinimisation}. However, the range of learning tasks that DP was applied to has widened and now includes nearly anything from the federated ML setting \cite{FederatedLearning_LearningFromMultipartyData} to recurrent language models \cite{DP_McMahan2018LearningDPLanguageModels}, and even GANs \cite{DP_GAN, DP_FunctionalMechanismForGANs}, with specific DP-GAN applications for generating time-series \cite{DP_GAN_forTimeSeries, time-GAN}, and tabular mixed feature datasets \cite{DP_data_withGANs}.


\subbold{Evaluation and Utility-Privacy Trade-Off of DP methods.}  The utility vs privacy trade-off is one of the most important topics in DP, partly due to the lack of formal utility-loss guarantees \cite{DP_CalibratingNoiseToSensitivity, jia2019memguard}.
Evaluation of privacy guarantees for DP is more established compared to some of the other defence methods. However, despite the various DP methods, and the provable upper bounds on their maximum privacy loss, there is still relatively little understanding of the trade-off between the size of the privacy budget $\epsilon$ and the utility of the resulting model. It is typical to select large values for $\epsilon$ to show reasonable utility scores \cite{DifferentialPrivacy_Jayaraman2019WhenRG, DifferentialPrivacy_Jayaraman2019EvaluatingDP}. Practically, \cite{DifferentialPrivacy_Jayaraman2019EvaluatingDP} finds that there is a huge gap between the guaranteed upper bounds on privacy loss, and the effective privacy loss that can be measured using inference attacks. Moreover, there is no agreed upon threshold for $\epsilon$, at which privacy guarantees are considered meaningless. The empirical assessment shows that for an acceptable utility level privacy guarantees are practically meaningless, although the observed level of leakage under the inference attacks is still low. Advancing further on DP under inference attacks, \cite{DP_ChallengingDP_nonInterractiveMechninsms} offers more empirical assessment of leakage under inference attacks, considering single and joint decoding (MIA, see Sec.~\ref{subsec:MIAs} for single data instance vs a subset of instances at a time), finding the joint decoding is more powerful, and offering a method to empirically choose the size of privacy budget $\epsilon$.

Some research has been conducted on eliminating the privacy-utility trade-off, replacing it with privacy-computational cost trade-off \cite{DP_LanguageModels_withoutLosingAccuracy}. They propose a stochastic gradient descent-based DP (Sec.~\ref{sec:defence_SGD}) for recurrent language models in a federated learning setting (Sec.~\ref{sec:defence_SGD}).

\subbold{Risks.} DP has been shown to be insecure under PIAs (see Sec.~\ref{subsec:PIAs}), because of the different types of data leakage considered by PIA and DP \cite{PropertyInferenceAttacks_Hacking_extractDataFromClassifiers}. Moreover, $(\epsilon, \delta)$-differential privacy retains the possibility of failures, i.e. a DP algorithm can in theory reveal the sensitive data it has been trained on. According to \cite{DP_PATE_and_noisySGD}, no mechanism has been proposed for detection and reporting of this kind of leakage.

For neural networks, two more recent approaches of implementing DP are particularly relevant, and the next two subsections are dedicated to these. Note that these are merely sub-classes of DP methods, and share general limitations and vulnerabilities of DP methods.



\subsubsection{DP: Private Aggregation of Teaching Ensembles \cut{(PATE)}}\label{subsec:PATE}
\textit{Private Aggregation of Teaching Ensembles (PATE)}, \cite{DP_PATE_Nissim07smoothsensitivity, FederatedLearning_Multiparty_PATEforClassifiers, FederatedLearning_LearningFromMultipartyData}, and its modification \textit{PATE-G},  \cite{PrivacyPreserving_Papernot2017SemisupervisedKT}, is a subset of differential privacy techniques based on the teacher-student approach, using ensemble methods (\cite{DP_EnsembleMethods_genericSurvey}) aggregation and some of the GAN-based architecture for PATE-G, \cite{GANs_NIPS2014_5ca3e9b1, GANs_Improved_NIPS2016_8a3363ab}. 

At training time an ensemble of teacher networks is trained on disjoint subsets of the training dataset with strong privacy guarantees, then the student network is used to aggregate the teacher network's knowledge in a noisy fashion, i.e. the student is black-box-querying the teacher ensemble, receiving the noisy labels. PATE methods train the student only on labelled training data, whilst PATE-G also uses the unlabelled data (via GANs or Virtual Adversarial training). At inference time, only the student model is used. Teacher models are never publicly shared, and student models never see the training data, thus the noisy aggregation of the teacher ensemble provides privacy guarantees \cite{DP_PATE_and_noisySGD}.

The scalability of PATE methods has been practically confirmed by \cite{DifferentialPrivacy_PATE} (on SVHN and the UCI Adult datasets). They further proposed to use concentrated noise (swapping Laplacian for Gaussian noise during aggregation) for further improvement of the teacher ensemble results, as well as withholding an answer to the student network at training time in the absence of teacher ensemble consensus. They report both high utility and privacy guarantees for $\epsilon < 1$.





\subbold{Applications} Theoretically, PATE can be universally applied to a variety of models. Although more classical works are applied to classifiers, \cite{DP_G-PATE_studentGenerator_Long2019ScalableDP, DP_PATE-GAN_Jordon2019PATEGANGS} focus on the data generation with DP guarantees. G-PATE \cite{DP_G-PATE_studentGenerator_Long2019ScalableDP}, 
trains a student-generator with an ensemble of teacher discriminators. Note, G-PATE and PATE-G should not be confused -- G-PATE is merely one of the PATE-G methods. PATE-GAN \cite{DP_PATE-GAN_Jordon2019PATEGANGS} trains a student classifier on synthetically generated data, using a noisy aggregation of the teacher-discriminator labels.

\subsubsection{DP: the Gradient Descent Perturbations}\label{sec:defence_SGD}

Neural network training relies on gradient descent, so adding noise is a popular technique for better generalization \cite{DP_SGD_Noise_Generalisation_DBLP:journals/corr/NeelakantanVLSK15, DP_SGD_NoiseGeneralisation_pmlr-v119-smith20a} and (with appropriate calibration) for ensuring differential privacy \cite{DP_CalibratingNoiseToSensitivity}. Since weight changes with respect to the training data occur through a gradient update, both gradient clipping and adding noise to gradient computations are valid privacy-preserving techniques 
\cite{DP_SGD_withDPUpdates, DifferentialPrivacy_EmpiricalRiskMinimisation, Privacy-preservingDeepLearning, DP_VariationalInference_basedOnGradientsPerturbation}.

More recent advances of the noisy SGD include extension with the moments accounting technique \cite{DifferentialPrivacy_DeepLearningWithDP}, a scalable and computationally efficient ``bolt-on'' output perturbation technique by \cite{DifferentialPrivacy_ScalableSGDAnalytics}, and DP-LSSGD \cite{DifferentialPrivacy_DP-LSSGD}, based on Laplacian smoothing SGD, that stabilizes the training of DP models, leading to better generalization and higher utility of the resulting DP models. Finally, adaptive allocation of the privacy budget at the iteration level \cite{DP_GradientDescentWithPrivacyBudget}, and \cite{DP_VarianceReducedSGD} applying the control variates technique \cite{DP_SGD_ControlVariates_1, DP_SGD_ControlVariates_2_NIPS2012_905056c1} to stochastic gradient descent update are both compatible with zCDP (see Sec.~\ref{subsec:DP} for more details on zCDP).

\subsection{Federated / Collaborative Learning}\label{sec:defence_federated}

Federated (or collaborative) Learning (FL) trains an ML model on a central server, across multiple decentralized databases, holding local data samples, without exchanging them directly \cite{Defences_federatedOptimisation, Defences_federated_improvingCommunication, Defences_federated_DeepLearning_ModelAveraging}, thus, potentially mitigating risks of the direct data leakage. \mcorr{It is considered a popular but not completely reliable defence against MIAs \cite{MIA_Rahman2018MembershipIA_againstDifferentiallyPrivate, MIA_Nasr2019ComprehensivePA}.}

\subbold{Surveys.} There are a number of surveys covering FL in general, \cite{AttacksOnFederated_Yang2019FederatedML, Survey_Federated_challengesMethodsFuture, Survey_Federated_openProblems, Survey_federatedLearningSystems}. We would like to refer the reader to \cite{Survey_FederatedLearning_threats} which focuses mainly on privacy concerns for FL. 

\subbold{FL vulnerability to privacy attacks.} FL is sometimes offered as a solution to the problem of balancing user data privacy requirements (such as GDPR \cite{GDPR}) with the benefits of learning from multiple data sources, \cite{AttacksOnFederated_Yang2019FederatedML}. However, FL does not provide foolproof privacy guarantees. Successful white-box MIAs have been performed by \cite{MIA_Nasr2019ComprehensivePA} against both centralised and federated learning, even for cases with well-generalised target models. These attacks leverage stochastic gradient descent (SGD) vulnerabilities; specifically they compute membership probability for each data point based on the gradient vector of all parameters with respect to this data point. Furthermore, \cite{Survey_FederatedLearning_threats} concludes that classic FL frameworks are often vulnerable to inference and poisoning attacks (Sec.~\ref{subsec:MIAs}
), also expressing concerns with the current methods of defences against these attacks for FL.

\subbold{Malicious servers} An alternative to a malicious user is a malicious server provider aiming to steal client's data. Recently \cite{AttacksOnFederated_Song2020AnalyzingUP} proposed the first ever attack from the perspective of such a malicious server.
It uses a GAN \cite{GANs_NIPS2014_5ca3e9b1, GANs_Improved_NIPS2016_8a3363ab} multi-task discriminator, designed to recover the category and the client identity of the input data. It is designed to run ``invisibly'' on a server leaving the clients unaware.

\subbold{Differential privacy for FL.} Efforts have been made to secure the classic FL framework relying on differential privacy \cite{FederatedLearning_Multiparty_PATEforClassifiers, DifferentialPrivacy_FederatedML_Geyer2017DifferentiallyPF, FederatedLearning_FastAndPrivateAlgosForFederated, DP_BayesLearningOnDistributedData, FederatedLearning_LearningFromMultipartyData, DP_connectingToSGD}. Some concerns remain on privacy-utility trade-offs \cite{DP_LanguageModels_withoutLosingAccuracy}, and property inference attacks for groups of records (rather than a single record) \cite{ExploitingUnintendedFeatureLeakage_inCollaborativeLearning}.

\subbold{Other defences for FL.} An important point raised in \cite{Survey_FederatedLearning_threats} is the lack of clarity on whether certain defences, such as adversarial defences, can be used for FL systems. 
A traditional defence for FL is homomorphic encryption (Sec.~\ref{subsec:HE}), used to mask the local gradient updates, either individually \cite{Defences_federated_HE_1, FederatedML_Liu2020ASF} or in batches for reducing computation costs \cite{Defences_federated_homomorphic}.

\subbold{Applications.} Federated learning is widely applied in applications involving the use of sensitive data, e.g., recommendation systems, mobile applications, transaction fraud detection, and healthcare \cite{AttacksOnFederated_Yang2019FederatedML, FederatedML_Liu2020ASF, PrivacyPreserving_PracticalSecureAggregation_Bonawitz2017PracticalSA, Survey_federatedLearningSystems}. Nevertheless, according to \cite{Survey_federatedLearningSystems}, there are not many FL applications that explicitly focus on privacy preservation. Still, there are some examples of privacy-preserving recommendation systems \cite{Defences_federated_privacyPreservingapplication, Defences_federated_privacyPreservingapplication_2}, that rely primarily on data encryption (see the next section) for their privacy guarantees.

\subsection{Operating on Encrypted Data}\label{subsec:HE}

Traditional encryption requires the sharing of the key amongst the parties involved, which interferes with individual privacy. However, \textit{Homomorphic Encryption (HE)} techniques allow any third party to operate on the encrypted data without decrypting it in advance, and, furthermore, \textit{Fully Homomorphic Encryption(FHE)}, \cite{HE_FHE_Gentry}, allows for any computable function to perform on the encrypted data \cite{Survey_HE}. 

\subbold{Surveys.} Homomorphic Encryption is a vast and well-established field, hence, for the sake of brevity, we refer the reader to the relevant surveys \cite{Survey_HE, Survey_HE_Brakerski2018FundamentalsOF}.

\subbold{Limitations.} Operating on encrypted data could alleviate privacy issues, but unfortunately its low efficiency often makes FHE impractical in the real world \cite{Defences_HE_practical}. However, newer advances, e.g., somewhat homomorphic encryption aim to improve efficiency -- refer to \cite{Defences_HE_Privacy_TaxonomyAndSurvey} for more details.





\subsection{Other Privacy-preserving ML}

Various other methodologies and applications exist for protecting against malicious leakage at model level. PRADA \cite{ModelExtraction_Defence_PRADA_againstDNNstealingAttacks} defends against model extraction attacks by flagging multiple queries made against a model when they deviate against general inference behaviour. 
Privacy preserving alternatives to SGD, introduced in \cite{PrivacyPreserving_PPPhong2018PrivacyPreservingDLActivationFunction} and \cite{PrivacyPreserving_Phong2019PrivacyPreservingDL_viaWeightTransmission}, apply to a case when multiple data owners wish to train models combining their data keeping individual privacy, based on sharing weight parameters instead of gradient updates. FPPDL \cite{PrivacyPreserving_Lyu2019TowardsFA_blockchain} is a decentralized privacy preserving framework based on Blockchain for decentralization, and differential privacy (DPGAN) along with a 3 layer onion encryption to facilitate fairness. VIPS \cite{Privacy-Preserving_VariationalBayes_inPrivacySetting} overcomes the high amount of additional noise needed to make variational Bayes privacy preserving by combining a moment accountant to get a tight bound on the privacy cost of multiple VB iterations. 














\section{Metrics}\label{sec:metrics} 

Assessment of the data leakage in trained machine learning models remains an open area of research. Measuring leakage is case-specific, as it depends on the data type and the type of malicious/involuntary leakage in question. Further, any knowledge about the exact type and architecture of the attack might be crucial for the defence.

\subbold{Assessing involuntary leakage} may be easy for some components e.g. overfitting via generalization error. Others, such as memorization and feature leakage, are harder to troubleshoot. An \textit{exposure} metric \cite{TheSecretSharer_EvaluatingUnintendedMemorisation, TheSecretSharer_MeasuringMemorisation_and_ExtractingSecrets}, explained in Sec.~\ref{sec:mem} estimates model memorization for text data (thus far, no extensions to other data types exist). 
An assessment proposed by \cite{DEJAVU} focuses on estimating memorization in the lower layers of convolutional neural networks. 

\subbold{Assessing data leakage via attacks} can be occasionally be straightforward, e.g., for MIAs the membership inference easily translates into the re-identification score \cite{MIA_Long2017TowardsMeasuringMembershipPrivacy, ReID_usingGenerativeModels}. \mcorr{Further, \cite{Metrics_SystematicEvaluationOfPrivacyRisksInML} introduces a \textit{privacy risk score} to measure an
individual sample’s likelihood of being a training member, to identify samples with high privacy risks under MIAs.}

\subbold{Assessing data leakage for defence purposes} Examples of this include Kolmogorov-Smirnov distance used for verifying forgetting in \cite{LiuHaveYouForgotten}, metrics proposed by \cite{MachineUnlearning_geopardisesThePrivacy} for assessing machine unlearning leakage under MIAs, as well as some work on estimating the Bayes risk of the system \cut{via universally consistent nearest neighbor (ML) rules} \cite{F-BLEAU-fast-black-box-leakage-estimation}, improving upon more na\"{i}ve min-entropy approaches. \mcorr{More specifically, \cite{F-BLEAU-fast-black-box-leakage-estimation} proposes a number of Bayesian metrics based on universally consistent nearest neighbor rules, from which metrics that converge the fastest should be selected. This provides an estimate of the Bayes risk of the model i.e., the error of the optimal classifier for predicting a sensitive attribute given an output observation from the model.}

\subbold{Learning metrics as a fairness constraint} Most literature in fair ML deals with learning fair classifiers\cut{ due to the prominence of classification as a learning task}. Most proposed methods treat solving for fairness based on the definition of fairness tailored to their specific objective. Of considerable importance are techniques such as\cut{ those proposed by} \cite{huang2019stable} which not only satisfy fairness constraints, but also tend to be stable towards adversarial attacks and variations in datasets during testing. Regression-based fairness techniques eliminate bias at training time by hand-crafting loss functions that conform to group, individual or hybrid fairness, although they have not received a lot of attention so far \cite{berk1706convex}.

\subbold{Metrics in Differential Privacy} In this setting, due to the 
provable privacy upper bounds, empirical assessment of both utility and quality guarantees is possible \cite{Defences_DiffPrivate_SyntheticData_Abay2019, DifferentialPrivacy_Jayaraman2019EvaluatingDP}. The Rényi Divergence can be used as a metric to bound any arbitrary privacy loss 
\cite{DifferentialPrivacy_Jayaraman2019EvaluatingDP}. The resulting Rényi differential privacy works by creating a bound on each individual moment of the privacy loss, leading to other variants of differential privacy, and to a more accurate numerical analysis of the privacy loss. A synthetic data generating deep learning model with privacy guarantees (DP-SYN) was proposed in \cite{Defences_DiffPrivate_SyntheticData_Abay2019}. Evaluation of DP-SYN was done using carefully crafted metrics based on ML (misclassification rate), statistics (Total Variation Distance \cite{metrics-total-var-dist} between the noisy and original marginals of the data distributions) and agreement rate (the percentage of records to which two classifiers assign the same prediction \cite{metrics-agreement-rate}). 

\subbold{Limitations.} First of all, there are a number of attacks/leakages that can be hard to trace. For instance, there is currently no single reliable way to verify how much of the training data is memorized by a GAN, or how much a property inference attack could infer even from sanitized data, since it would change, depending on the design of the attack and the type of the data in question.

Secondly, there is no universal robust framework for detecting and reporting model plainly revealing the sensitive data (more likely for predictive or generative models), \cite{DP_PATE_and_noisySGD}, and although it does not necessarily seem like a big issue at first glance, it does impede open access trained model sharing in a commercial setting, as companies will require guarantees on the privacy of their data.






\section{Applications}\label{sec:apps}
\subsection{Data as a Service (DaaS)}\label{sec:daas}
Data as a Service offers an appealing solution to limited data availability in both data-driven research and data-intensive commercial applications, given sufficient privacy guarantees. However, current proposed implementations, e.g. \cite{App_DaaS_GPUimplementation}, provide no leakage assessment. 
Significant efforts are afoot to create national research infrastructures across \cut{the United Kingdom and across }the world to support data-driven research\cut{ knowledge discovery}
. Organizations such as Health Data Research UK and Research Data Scotland design services for identification of health research datasets, their description, permissions, and accessibility.

Federated access models are favoured by data holders, with data scientists invited to access data within Trusted Research Environments, for example Data Safe Havens. However, with the intellectual and economic benefits of more access to data comes an escalating risk of data leakage, driving persistent privacy concerns. 

The current official protocols in the UK, rely on statistical disclosure control \cite{App_DaaS_ONS_GuidanceOnIntruderTesting, App_DaaS_NHSStatisticalDisclosure}, data pseudoanonymization \cite{App_DaaS_EUCybersecurity_PseudoAnonymisation}, and true anonymization \cite{App_DaaS_ICO_manualOnAnonymisation}, since fewer legal restrictions apply to anonymized data. 
However, from a legal perspective, anonymized data is a gray area \cite{App_DaaS_TheyWhoShallNotBeIdentified}.
In fact, regulations such as Data Protection Directive (1995) \cite{App_DaaS_DataProtectionAct_EU}, Data Protection Act 1998 (DPA) \cite{App_DaaS_DataProtectionAct}, and GDPR \cite{GDPR}, do not require strictly risk-free data protection, but the risk of re-identification should be mitigated to the extent when it is remote. GDPR does not apply to truly anonymized data either -- Recital 26 defines the anonymous information, as ``\textit{ information which does not relate to an identified or identifiable natural person or to personal data rendered anonymous in such a manner that the data subject is not or no longer identifiable}'', \cite{App_DaaS_GDPR_Recital26}. However it applies to pseudo- and non- anonymized data, often more useful in practice, and preferable for statistical analysis and ML applications in DaaS setting.

If DaaS using linked and unconsented public data, under the GDPR safeguards and standards for privacy is to continue, techniques to mitigate data linkage are imperative. Hence custodians considering DaaS, especially with sensitive datasets, have a difficult dual duty to both respect public privacy, and to foster public benefit through research.

From an ML perspective, this results in a growing demand for the \cut{implementation of }reliable checks on models exported from DaaS facilities\cut{. This requires furthering our understanding and control over involuntary data leakage, and progressively}\mcorr{, including} more reliable methods of defence from privacy attacks, such as MIAs and PIAs (Sec.~\ref{sec:leakage}, \ref{subsec:MIAs} and \ref{subsec:PIAs}).





\subsection{ML Models as a Service (MLaaS)} \label{sec:mlaas}

Machine Learning as a Service (MLaaS) represents an extended privacy risk, further to that posed by DaaS. The development of ML models can risk perpetuating bias, state intrusion, inequalities, and erosion of privacy. Whilst the separation of source data from MLaaS could ameliorate data leakage concerns, the outputs, decisions, and unintended applications of MLaaS complicate tracing potential leakage. 

Certain settings of MLaaS, including federated learning, can be vulnerable to inference type attacks, e.g., MIAs \cite{MIA_DemystifyingMIAsMLasService, DefencesDataLevel_Wang2020MIASecED}, with defence mechanisms shown to mitigate those risks explored mainly for classification \cite{DefencesDataLevel_Wang2020MIASecED, DifferentialPrivacy_ModelPublishing}.
Moreover, \cite{Reconstruct_inverseShadowModel_beatsAmazonMLaaS} showed that Amazon Rekognition, a commercial MLaaS API, can be vulnerable to model inversion attacks. Research in MLaaS data safety remains important for understanding the risks posed by models as they\cut{ are deployed, trained, and} evolve on exposure to new data.
Presently\cut{ a number of} providers such as Amazon ML \cite{App_MaaS_amazon}, Google Cloud \cite{App_MaaS_GoogleCloud}, and IBM \cite{App_MaaS_IBM} are providing MLaaS for public and commercial use. \cut{The implications of open-access-sharing ML models trained on sensitive data and current defences for these are covered in Sections~\ref{sec:leakage} -- \ref{sec:defences_model} of this survey.}
Sections~\ref{sec:leakage} -- \ref{sec:defences_model} cover the implications of open-access-sharing ML models trained on sensitive data and current defences. 






\subsection{ML models in Mobile Applications} \label{sec:mobile}
ML methods are often used to support mobile applications. Thus, privacy attacks, e.g., MIAs \cite{MIA_ML}, attribute inference attacks, and PIAs (Sec.~\ref{subsec:MIAs} and \ref{subsec:PIAs}) are a possibility. Sensitive information could be anything from the full user profile or location \cite{MIA_OnAggregateLocationData, UnderTheHoodOfMIAonAggregateLocationTS} to gender and sexual orientation \cite{AttriGuard}. 

Federated learning (see Sec.~\ref{sec:defence_federated} for more details on risks and defences) appeals in this context as means of privacy protection. Although some research for protecting mobile users specifically exist, e.g., \cite{AttriGuard} and \cite{PrivacyPreserving_PracticalSecureAggregation_Bonawitz2017PracticalSA}, this field is still somewhat in its adolescence.
 

\section{Challenges and Opportunities} \label{sec:challenges}

Our findings thus far can be summarized as follows:


\subbold{Attacks} are not evenly explored across different data types or tasks. For instance, MIAs (Sec.~\ref{subsec:MIAs}) are not well investigated for tasks such as regression or segmentation, MEAs (Sec.~\ref{sec:MEA}) have not been verified for generative models, and PIAs have only been applied to classification. 
This points to the need to uniformly probe weaknesses of leakage across several tasks and data types via advancements in attacks.

\subbold{Defences at the data level} lie in between data being potentially anonymized (or sanitised, obfuscated) to the point where they are no longer useful, and data being likely re-identifiable through inference attacks. Replacing real personal data with synthetic data could be a promising direction, albeit they remain vulnerable to property and attribute inference attacks \cite{Defences_SyntheticData_PrivacyMirage_2020}. 
Data privacy can be largely contextual, i.e., in certain situations a publicly accessible dataset can potentially enable recovering individuals' identities, when combined with other public datasets.


\subbold{Defences via model} have not been evenly explored across different tasks, data, and attacks/leakage, and may often work only for specific settings.
For example, adversarial defences are mainly explored for MIA-type attacks, DP-based defences may not universally succeed against MIAs, and the privacy guarantees of classic FL may not be as strong as we desire.
Homomorphic Encryption is a promising direction for privacy-preserving FL; however, its practical implementation is not straightforward and requires compute power and a homogenous setup across all parties involved. We remain in need of computationally efficient defences, which can offer a wide range of privacy guarantees.  

\subbold{Detection and Assessment of Leakage and Tools} Furthermore, uniform mechanisms for reporting data leakage are lacking. For instance, in a DaaS scenario a malicious user could potentially encode sensitive data within the NN model weights -- yet a check / mechanism to reliably detect even such a simple form of leakage is lacking. We find that established  and universally applicable software packages based on already existing research are lacking. This results in an opportunity to develop  mechanisms for transparent reporting and create robust software that can help bridge the gap between theory and practical utility. 












\section{Conclusion}\label{sec:conclusion}
While data leakage research is not new, the field is ever-evolving due to the dynamic (and rapid) nature of machine learning development. New privacy risks and attacks arise, prompting new efforts to protect against them, resulting in a constant adversarial game. This survey unifies and summarizes current research into inference-time information leakage in ML, both involuntary and malicious, as well as the means currently available to measure and prevent such leakage. This results in a rich comprehensive taxonomy of the broad field of privacy in ML.

We find, first of all, that understanding of data leakage, its causes and implications, is unexplored and our hope is that this survey will positively contribute towards furthering our appreciation of data leakage. Our survey reveals opportunities to improve the means to measure, detect and report sensitive data leakage. Secondly, privacy attacks exploration has been uneven in its coverage of ML tasks and architectures, data types, and attack structures. Finally, we find that most available defences are case-specific, and scaling to larger datasets with performance guarantees remains a challenge. Overall this indicates that leakage, privacy, and the necessary defenses, remain areas which are fertile for further research and development.


\vspace{-0.25cm}
\section*{Acknowledgment}
This work is supported by iCAIRD, funded by Innovate UK, UK Research and Innovation (UKRI)[104690]. S.A. Tsaftaris acknowledges support by Canon Medical / Royal Academy of Engineering Research Chair, Grant RCSRF1819\textbackslash8\textbackslash25.
\vspace{-0.25cm}


\ifCLASSOPTIONcaptionsoff
  \newpage
\fi



\bibliographystyle{IEEEtran}
\bibliography{IEEEexample_redacted}

\vspace{1.3mm}
\begin{wrapfigure}[8]{L}{23mm}
    \vspace{-3mm}
    \includegraphics[width=0.8in,keepaspectratio]{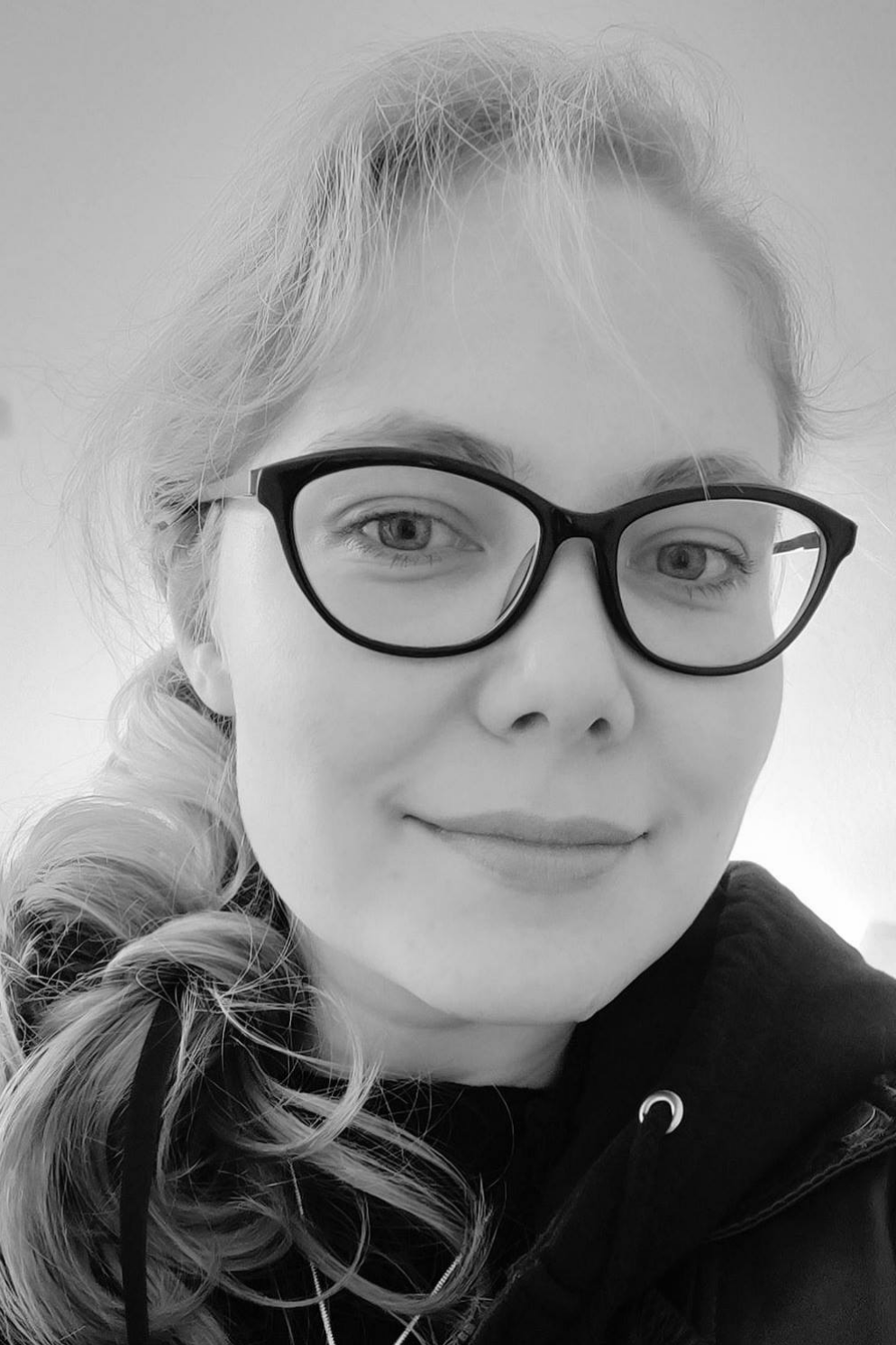}
\end{wrapfigure}
\textbf{Dr Marija Jegorova} is a postdoc at Facebook AI (London) and a former postdoc at School of Engineering, the University of Edinburgh. Her PhD focused on realistic artificial data synthesis for ML applications to robotics. Current research interests include semi-supervised learning and video emotion recognition.


\begin{wrapfigure}[8]{L}{23mm}
    \vspace{-3mm}
    \includegraphics[width=0.8in,keepaspectratio]{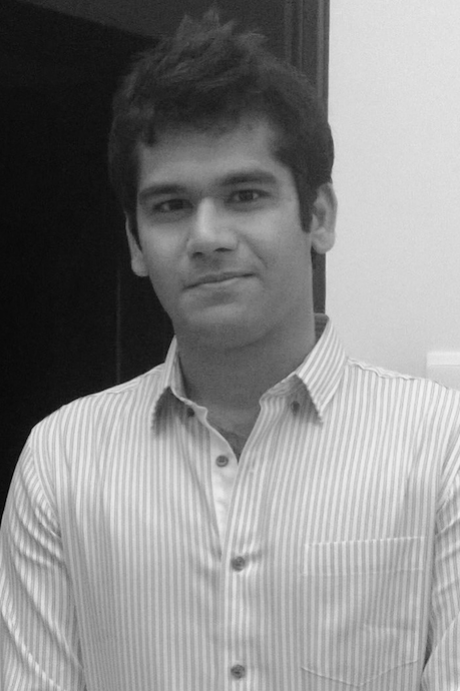}
\end{wrapfigure}
\textbf{Dr Chaitanya Kaul} is a Research Associate in the School of Computing Science at the University of Glasgow. His research interests include medical image analysis, attention models for deep neural networks, generative modelling, and 3D shape analysis. He works on private and explainable machine learning.


\begin{wrapfigure}[8]{L}{23mm}
    \vspace{-3mm}
    \includegraphics[width=0.8in,keepaspectratio]{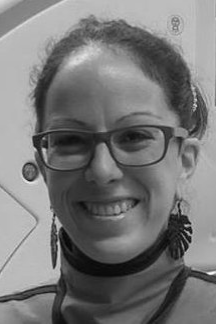}
\end{wrapfigure}
\textbf{Dr Alison~Q.~O'Neil} is a Senior Scientist in the AI Research Team at Canon Medical Research Europe and Honorary Research Fellow at the University of Edinburgh. She leads a team working on machine learning for industrial healthcare applications, such as medical imaging, NLP, and electronic health records.


\begin{wrapfigure}[8]{L}{23mm}
    \vspace{-3mm}
    \includegraphics[width=0.8in,keepaspectratio]{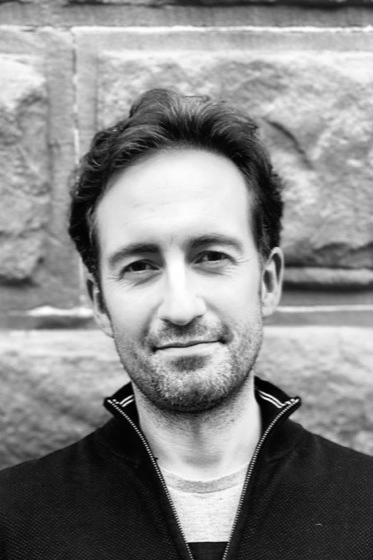}
\end{wrapfigure}
\textbf{Dr Charlie Mayor} is a Manager at Safe Havens Scotland, NHS Greater Glasgow and Clyde. Current research interests include development and deployment of privacy-preserving methodologies and safety checks in order to meet the potential future requirements for the safeguards of the Safe Havens.



\begin{wrapfigure}[8]{L}{23mm}
    \vspace{-3mm}
    \includegraphics[width=0.8in,keepaspectratio]{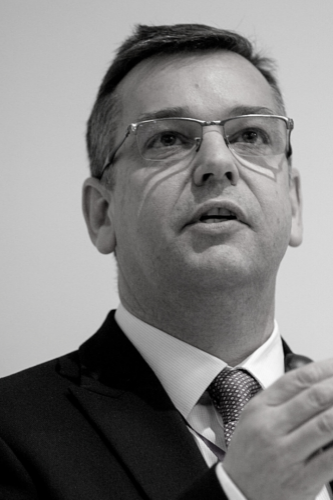}
\end{wrapfigure}
\textbf{Dr Alexander Weir} is a Senior Technical Manager at Canon Medical Research (Europe). He leads specialist project research teams developing new medical systems for patient medical data management, deploying and developing state of the art applications to industrial healthcare technology.


\begin{wrapfigure}[8]{L}{23mm}
    \vspace{-3mm}
    \includegraphics[width=0.8in,keepaspectratio]{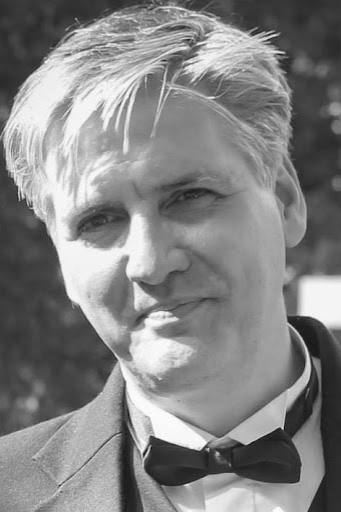}
\end{wrapfigure}
\textbf{Prof. Roderick Murray-Smith} is a Professor of Computing Science at the University of Glasgow, in the \textit{Inference, Dynamics and Interaction} group. He works in the overlap between machine learning, interaction design and control theory. His more recent interests include quantum imaging, multimodal sensor-based interaction with mobile devices, mobile spatial interaction, Brain-Computer interaction, and nonparametric machine learning.


\begin{wrapfigure}[8]{L}{23mm}
    \vspace{-3mm}
    \includegraphics[width=0.8in,keepaspectratio]{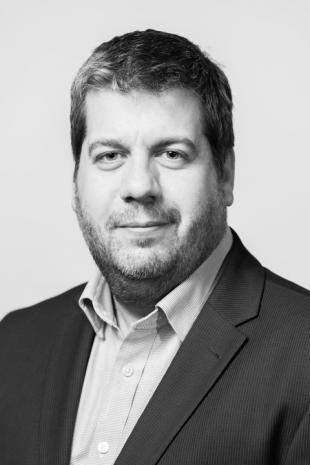}
\end{wrapfigure}
\textbf{Prof. Sotirios A. Tsaftaris} is a Chair in Machine Learning and Computer Vision at the University of Edinburgh, and holds the Canon Medical/Royal Academy of Engineering Research Chair in Healthcare AI. His research interests include machine learning, computer vision, distributed computing and applications in healthcare and other domains. 

\end{document}